\documentclass[10pt,twocolumn,letterpaper]{article}

\usepackage[pagenumbers]{cvpr}
\usepackage{microtype}
\usepackage{xspace}

\newcommand{\method}{PhysMend\xspace}
\newcommand{\papertitle}{\method: Verification-Guided Program Refinement\\for Physics-Aware Video Generation}
\newcommand{\physicsiq}{Physics-IQ\xspace}
\newcommand{\phygen}{PhyGenBench\xspace}
\newcommand{\vbench}{VBench-2.0\xspace}

\newcommand{\samthree}{SAM~3\xspace}
\newcommand{\depthanything}{Depth Anything~V2\xspace}
\newcommand{\wanvace}{Wan-VACE\xspace}
\newcommand{\waniiv}{Wan2.2-I2V-A14B\xspace}
\newcommand{\wantiiv}{Wan2.2-TI2V-5B\xspace}
\newcommand{\cogvideoxi}{CogVideoX-5B-I2V\xspace}
\newcommand{\ltxvideo}{LTX-Video\xspace}
\newcommand{\svdxt}{SVD-XT\xspace}
\newcommand{\sgiivid}{SG-I2V\xspace}
\newcommand{\gowithflow}{Go-with-the-Flow\xspace}
\newcommand{\gptfouro}{GPT-4o\xspace}
\newcommand{\llavainterleave}{LLaVA-Interleave\xspace}
\newcommand{\genthree}{Gen-3\xspace}

\usepackage{algorithm}
\usepackage{algpseudocode}
\usepackage{tikz}
\usetikzlibrary{arrows.meta,positioning,calc}
\newsavebox{\rankscorebox}
\DeclareRobustCommand{\rankhighlight}[2]{%
  \begingroup
  \sbox{\rankscorebox}{#2}%
  \tikz[overlay,baseline=0pt]{%
    \fill[#1,opacity=0.15]
      (-2pt,{-0.3ex-0.5pt}) rectangle
      ({\wd\rankscorebox+2pt},{1.35ex+0.5pt});}%
  \usebox{\rankscorebox}%
  \endgroup}
\DeclareRobustCommand{\best}[1]{\rankhighlight{red}{#1}}
\DeclareRobustCommand{\second}[1]{\rankhighlight{blue}{#1}}

\usepackage{tabularx}

\definecolor{cvprblue}{rgb}{0.21,0.49,0.74}
\usepackage[breaklinks,colorlinks,allcolors=cvprblue]{hyperref}
\usepackage{refcount}

\def\confName{CVPR}
\def\confYear{2027}

\title{\papertitle}

\ifdefined\standalonesupplement
  \usepackage{xr}
  \providecommand{\fullauxfile}{main}
  \IfFileExists{\fullauxfile.aux}
    {\externaldocument[main-]{\fullauxfile}}
    {\externaldocument[main-]{main}}
  \newcommand{\mainref}[1]{\ref*{main-#1}}
\else
  \newcommand{\mainref}[1]{\ref*{#1}}
\fi
\ifdefined\standalonesupplement
  \newcommand{\mainfloatref}[1]{\ref*{main-#1}}
\else
  \newcommand{\mainfloatref}[1]{\ref{#1}}
\fi
\newcommand{\inmainpaper}{ in the main paper}
\makeatletter
\newcommand{\supptitle}[1]{{\let\texorpdfstring\@secondoftwo\nameref*{#1}}}
\makeatother

\newcommand{\continuedcounters}{figure,table,equation,algorithm,footnote}
\makeatletter
\newcommand{\recordmaincounters}{%
  \@for\@tempa:=\continuedcounters\do{%
    \immediate\write\@auxout{\string\newlabel{maincount:\@tempa}%
      {{\the\csname c@\@tempa\endcsname}{}{}{}{}}}}}
\newcommand{\continuemaincounters}{%
  \@for\@tempa:=\continuedcounters\do{%
    \refused{main-maincount:\@tempa}%
    \setcounter{\@tempa}{\getrefnumber{main-maincount:\@tempa}}}}
\makeatother

\author{%
  Meilong Xu$^{1*}$\quad Ziyi Zhao$^{2}$\quad Bhagyashree Puranik$^{2}$\quad Shayda Moezzi$^{3}$\\
  Yixiao Zhang$^{2}$\quad Kushan Thakkar$^{2}$\\[0.4em]
  {\normalsize $^{1}$Stony Brook University\quad $^{2}$Amazon\quad $^{3}$Northeastern University}\\
  {\small\href{mailto:meixu@cs.stonybrook.edu}{\tt meixu@cs.stonybrook.edu}}%
}

\newcommand{\Suppref}[1]{The supplementary material}
\newcommand{\suppref}[1]{the supplementary material}

\ifdefined\standalonesupplement
  \makeatletter
  \AtBeginDocument{%
    \def\hyper@natlinkstart#1{}%
    \def\hyper@natlinkend{}%
    \def\hyper@natlinkbreak#1#2{#1}}
  \makeatother
\fi

\begin{document}
\ifdefined\standalonesupplement
\else
  \twocolumn[{%
    \renewcommand\twocolumn[1][]{#1}%
    \maketitle
    \begin{center}
      \captionsetup{type=figure}
      \includegraphics[width=\textwidth]{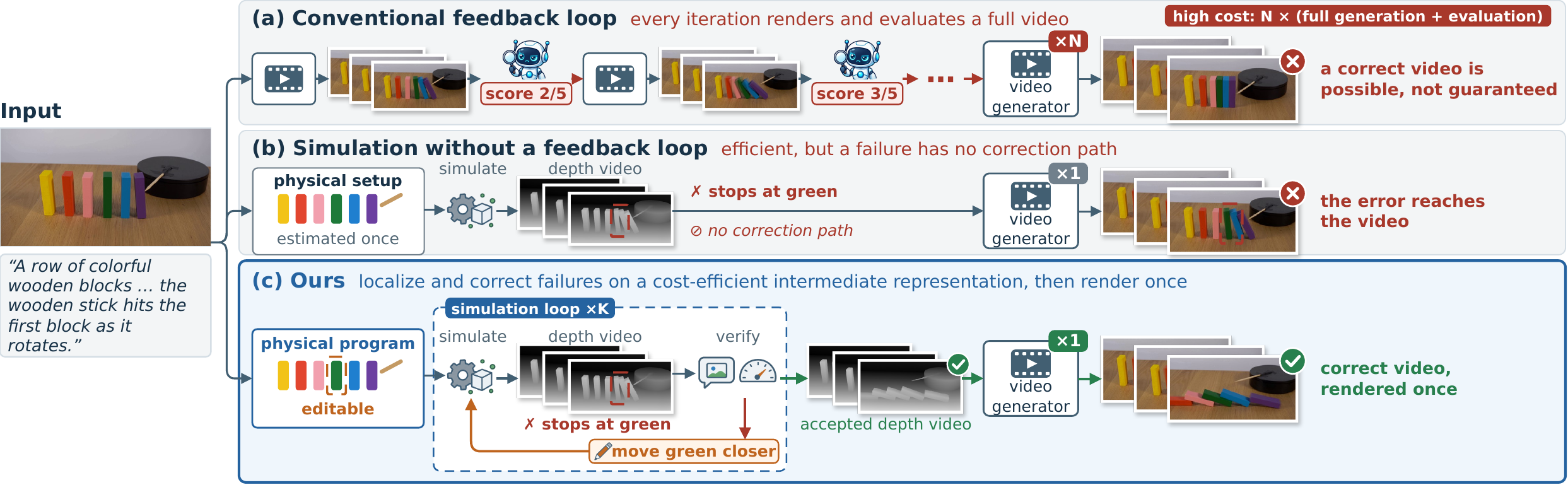}
      \captionof{figure}{\textbf{Comparison of feedback and correction strategies for video generation.} (a) Conventional feedback loops are costly because every iteration renders and evaluates a full video, and they may still fail. (b) Simulation without a feedback loop is efficient but has no correction path, so a simulation error reaches the video. (c) \method localizes each failure on a cost-efficient intermediate representation, corrects it with a targeted edit, and renders the full video only once.}
      \label{fig:teaser}
    \end{center}
  }]
  {\renewcommand{\thefootnote}{\fnsymbol{footnote}}%
   \footnotetext[1]{Work done during the internship at Amazon.}}
\fi

\ifdefined\standalonesupplement
\else
\begin{abstract}
Video generative models synthesize realistic scenes, yet their interactions often violate physics. A ball dropped onto sand should displace grains and settle into a crater, whereas a generated video may show it passing through the surface or leaving the sand flat. Post-hoc methods typically score generated videos with a vision-language model (VLM), but such scores offer limited guidance on which object or parameter to modify, and each correction costs another synthesis. Simulation-based pipelines instead run fixed stages before synthesis, so simulation errors reach the video. To address these limitations, we present \method, a training-free framework for verification-guided program refinement. \method describes the target event as an editable physical program of grounded objects, events, and required relations, paired with tunable execution parameters. 
Our key insight is to verify and repair the event on its simulated rollout and on the depth video that conditions the generator (depth control) rather than on generated videos, enabling quantitative violation measurement, targeted editing, and fast re-verification before synthesis. A hybrid verifier evaluates semantic event consistency, physical plausibility, solver stability, and visibility in the depth control. VLM judgments are confined to event semantics and cannot override failed physical measurements, which limits the impact of VLM hallucinations. Each failed check localizes the program fields eligible for repair, and its measured violation scales a bounded update of their parameters.
Only accepted depth control reaches the generator, which runs once per input. Experiments on the commonly used benchmarks \physicsiq, \phygen, and \vbench demonstrate substantial gains across different generator backbones; for example, \method raises the overall \physicsiq score of \wanvace from 27.33 to 39.40.
\end{abstract}

\begin{figure*}[!t]
  \centering
  \includegraphics[width=\textwidth]{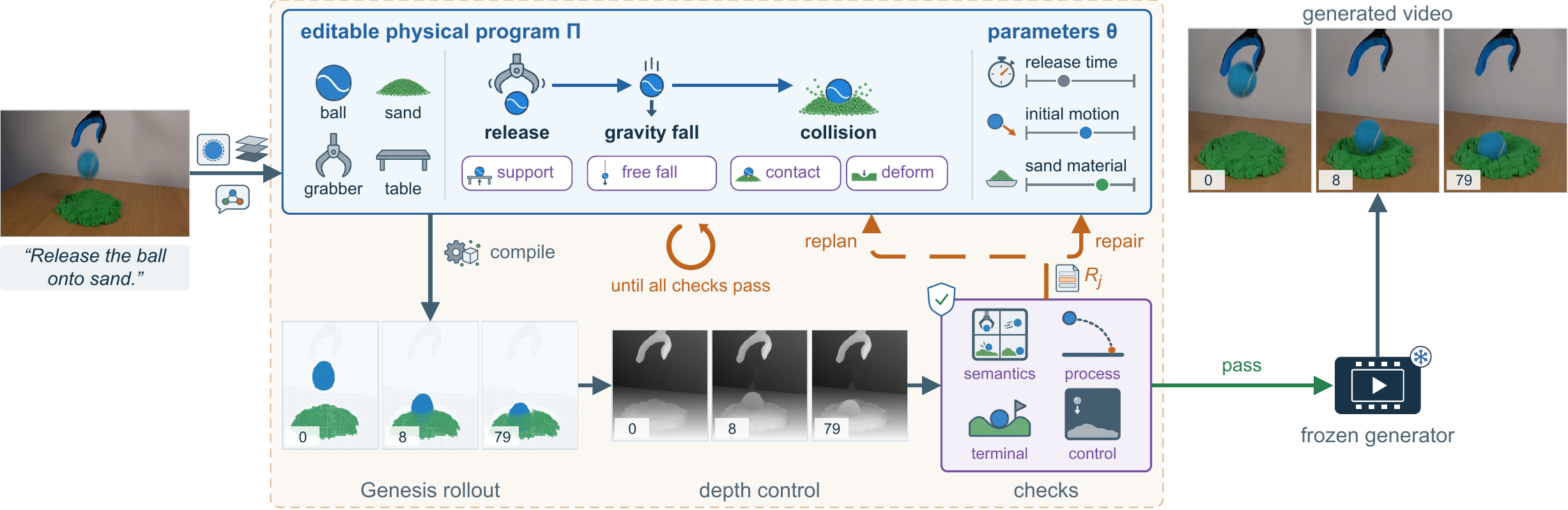}
  \caption{\textbf{Overview of \method.} Masks, depth, and the VLM turn the input into an editable physical program $\Pi$, compiled with parameters $\boldsymbol{\theta}$ and executed in Genesis~\cite{Genesis}. For the ball released onto sand, $\Pi$ lists the grabber, ball, sand, and table, and it chains release, fall, and collision, where the collision must contact and deform the sand. Its parameters $\boldsymbol{\theta}$ include the release time, initial motion, and sand material. Inside the shaded loop, checks examine the rollout and its depth control. Accepted depth control conditions a frozen generator once, whereas a failed check's record $R_j$ routes the loop (dashed) to replan $\Pi$ or repair $\boldsymbol{\theta}$. Frame indices are local to each sequence.}
  \label{fig:overview}
\end{figure*}

\section{Introduction}
\label{sec:introduction}
Video generative models have advanced rapidly. Closed-source systems (Sora~\cite{brooks2024video}, Kling~\cite{kuaishou2024kling}, \mbox{\genthree}~\cite{runway2024gen3alpha}, Seedance~\cite{seed2026one}) and open-source models (Wan~\cite{wan2025wan}, HunyuanVideo~\cite{kong2024hunyuanvideo}, CogVideoX~\cite{yang2025cogvideox}, MiniMax~H3~\cite{minimax2026h3}) synthesize realistic human activities, natural landscapes, and cinematic scenes. Despite this progress, they struggle to reproduce physically plausible object interactions and motion, such as collisions, contact, support, and trajectory changes, as recent analyses~\cite{kang2024far,thozhiyoor2026objects} and benchmarks such as \physicsiq, \phygen, VideoPhy-2, PhyWorldBench, and PVBench~\cite{motamed2026physics,meng2024towards,bansal2026videophy,gu2026phyworldbench,chen2026pvbench} report. A ball dropped onto sand must displace grains, form a crater, and come to rest, which is hard to reproduce because sand supports load yet flows under impact. A generated video can look realistic per frame while the ball passes through the surface or the sand stays flat, and poured liquid or falling silk fails similarly. We study physics-aware video generation from an initial image and an event prompt, where material properties, contact conditions, and interaction parameters are unobserved.

Existing methods act either after or before synthesis (Fig.~\ref{fig:teaser}). After synthesis, PhyT2V~\cite{xue2025phyt2v} refines its prompt from the generated video, and NEWTON~\cite{feng2026newton} uses a trained planner to select tools and revise conditioning. This feedback observes the output but relies mostly on VLM verdicts such as ``the contact looks wrong'' or scalar scores, which offer limited guidance on which object or parameter to modify, and each revision costs another synthesis call. Before synthesis, PhysGen~\cite{liu2024physgen} and PhysMotion~\cite{tan2026physmotion} supply simulated motion in fixed stages, so simulation errors reach the video.

Recent methods add verification and repair before synthesis~\cite{li2026videococo,huang2025planning,li2026physagent,xu2026veriphy}, but their repairs come from retries, execution errors, vision-language diagnoses, or prompt edits rather than from measured violations. 
A check reporting that the falling ball missed the sand does not specify whether to adjust its placement, release timing, or initial motion, or by what magnitude. Each edit must be rechecked, since it can break a requirement that already held. A valid rollout also helps only if projection or occlusion does not hide the contact from the generator. Refinement before synthesis must therefore localize what to change and by how much, recheck the event after each edit, and verify what the generator receives.

We meet these requirements with \method, a training-free framework for \emph{verification-guided program refinement} (Fig.~\ref{fig:overview}). Its key insight is to verify and repair the event on intermediate representations, namely its simulated rollout and the depth control rendered from it, rather than on generated videos. A prompt can describe an action such as releasing a ball onto sand, but not the physical setup and parameters needed to simulate it, such as object materials and supports or the ball's initial motion and release time. \method makes these details explicit in an editable physical program $\Pi$ of grounded objects, events, and required relations, compiled with tunable execution parameters $\boldsymbol\theta$ and simulated in Genesis~\cite{Genesis}. Each rollout and its depth control go through a hybrid verification mechanism that covers event semantics, the physical process, the terminal state, solver stability, and depth-control visibility.

Each failed check returns a record naming the affected objects, the checked interval, the violation size, and the editable fields. The record routes structural errors to replanning of $\Pi$ and measurable violations to bounded updates of $\boldsymbol\theta$ scaled by their size. Every edit is rechecked in simulation, so the program is refined by its own check results, and a rejected edit costs a simulation pass, not another video.

Our contributions are as follows:
\begin{itemize}
  \item We present \method, a training-free framework that turns an image and an event prompt into an editable physical program that can be checked on its rollout and depth control and edited before a single video synthesis.
  \item We build a hybrid verification mechanism that checks event semantics, physical processes, terminal states, solver stability, and depth-control visibility, and restricts VLM judgments to event semantics to limit the impact of VLM hallucinations on rollout acceptance.
  \item We design a verification-guided repair mechanism in which each failed check identifies the program fields eligible for repair, its measured violation scales a bounded update of their parameters, and each revision is rechecked and kept or rolled back within a fixed simulation budget.
\end{itemize}

Extensive experiments on \physicsiq~\cite{motamed2026physics}, \phygen~\cite{meng2024towards}, and \vbench~\cite{zheng2025vbench} show consistent improvements across backbones, including \wanvace~\cite{wan2025wan,jiang2025vace}, Cosmos3-Nano~\cite{agarwal2026cosmos}, and OmniWeaving~\cite{pan2026omniweaving}. Among the compared methods, \method achieves state-of-the-art performance on both \physicsiq and \phygen and remains comparable to state-of-the-art methods on \vbench.

\section{Related Work}
\label{sec:related}
\subsection{Physics-Aware Video Generation}
Video diffusion models extend image diffusion to temporal sequences~\cite{ho2020denoising, ho2022video}, with later architectures improving visual fidelity and enabling text- or image-conditioned synthesis~\cite{blattmann2023stable, wan2025wan, yang2025cogvideox}, yet physical interactions remain challenging~\cite{meng2024towards, motamed2026physics}. Training-based methods inject physical conditions~\cite{pathak2026physvid, wang2026wisa, zhang2025think}, transfer relational priors from foundation models~\cite{kim2026tempered, xiong2026physalign, zhang2026videorepa}, optimize physical rewards and preferences~\cite{chen2026hierarchical, deng2026ns, zhang2026physrvg, zhao2026phyworld}, or learn neural dynamics of forces, trajectories, and latent states~\cite{shen2026phantom, wang2026physctrl, yuan2026newtongen}.

Simulation provides a complementary, training-free motion prior. MotionCraft, PhysGen, and PhysMotion animate images with simulated dynamics~\cite{liu2024physgen, montanaro2024motioncraft, tan2026physmotion}, while PhysGen3D reconstructs interactive 3D scenes~\cite{chen2025physgen3d}. RealWonder converts physical actions into optical flow and visual guidance~\cite{liu2026realwonder}, and PSIVG reconstructs generated templates for simulation-guided synthesis~\cite{foo2026physical}. For longer horizons, Fluid-Gen-Zero plans clips from fluid rollouts to guide denoising~\cite{huang2026fluid}, whereas TourPhysics keeps simulated transitions fixed across observation retries~\cite{zhang2026tourphysics}. In contrast, \method verifies and repairs both the simulated rollout and its depth control prior to synthesis.

\subsection{Planning and Verification for Video Generation}
Pre-generation planning structures the target event before synthesis: VideoDirectorGPT decomposes prompts into layout plans~\cite{lin2023videodirectorgpt}, VLIPP and CausalMotion plan physical motion~\cite{yang2025vlipp, zhuang2026causalmotion}, and CoECT and its extension model causal event chains~\cite{wang2026physically, wang2026chain}. PhysPlan grounds state changes in the scene, validates typed graph edits, and guides denoising~\cite{nguyen2026physplan}, whereas \method simulates the compiled program and measures the rollout and the visibility of its depth control. Post-generation feedback instead revises outputs after synthesis by updating prompts from generated videos~\cite{long2026vista, xue2025phyt2v}, adjusting initial noise via flow residuals in Proprio~\cite{hassan2026proprio}, coordinating tools with learned policies in NEWTON and VideoGen-Agent~\cite{feng2026newton, li2026videogen}, or optimizing inference policies in SCOPE~\cite{jiang2026scope}. Rather than tuning a shared policy or regenerating videos, \method refines an event-specific physical program before a single generator call.

Recent works also verify intermediate plans before synthesis. SketchVerify evaluates sketches of candidate motion plans and retries failed ones~\cite{huang2025planning}, and VideoCoCo repairs Blender scripts via execution feedback before translating drafts with a trained editor~\cite{li2026videococo}, though syntactic executability alone does not guarantee the intended event. PhysAgent edits physical programs from vision-language diagnoses of reconstructions and rollouts to guide generation with optical flow~\cite{li2026physagent}. VeriPhy revises scenes before synthesis, but its evaluated video-feedback loop updates the prompt while the control stays fixed~\cite{xu2026veriphy}. Unlike repairs driven by retries, execution feedback, or vision-language diagnoses, \method maps each measured violation to a bounded parameter update over localized fields and rechecks both the rollout and its depth control before synthesis.

\section{Method}
\label{sec:method}

\begin{figure*}[!t]
  \centering
  \includegraphics[width=\textwidth]{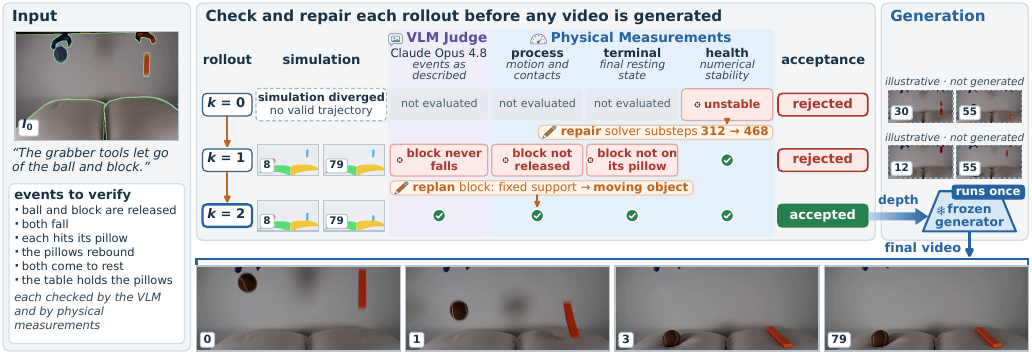}
  \caption{\textbf{Quantified verification and repair on one representative example from \physicsiq~\cite{motamed2026physics}.} Each rollout is simulated, and the VLM and the physical measurements check every event. At $k=0$, the simulation is numerically unstable, so no other check is evaluated, and parameter repair raises the solver substeps. At $k=1$, the program treats the hanging block as a static fixture. The VLM and the physical measurements fail on the block's events, and replanning makes the block a moving object. At $k=2$, every check passes, and the frozen generator runs once.}
  \label{fig:verification}
\end{figure*}

\subsection{Preliminaries}
\label{sec:preliminaries}

Genesis~\cite{Genesis} simulates gravity and contact with rigid primitives for rigid bodies, fixed geometry for supports and containers, the Material Point Method (MPM)~\cite{sulsky1994particle} for deformable and granular materials, and Smoothed Particle Hydrodynamics (SPH)~\cite{gingold1977smoothed} for liquids.

Control-conditioned video generators such as VACE~\cite{jiang2025vace} accept a dense control video in addition to the text prompt and initial frame. When this control is a depth sequence, it fixes scene geometry and object motion while the model renders appearance, which is how \method passes simulated motion to the generator.

\subsection{Overview}
\label{sec:formulation}

Given an initial image $I_0$ and a text prompt $p$ describing a target physical event (Fig.~\ref{fig:overview}), \method represents the event as an editable physical program $\Pi$ compiled with execution parameters $\boldsymbol\theta$. Simulating them at attempt $k$ yields a \emph{rollout} $\tau^{(k)}$, a sequence of simulated object states, whose projection into the input camera forms the \emph{depth control} $D_{1:T_g}$ of a depth-conditioned generator $G_\phi$ with frozen weights $\phi$ and $T_g$ output frames. Because simulation reaches the video only through this control, \method verifies and repairs the event on $\tau^{(k)}$ and $D_{1:T_g}$ before the generator runs.

Alg.~\ref{alg:main} summarizes the procedure: \method builds and executes the program (Sec.~\ref{sec:program}), checks the rollout and its depth control (Sec.~\ref{sec:verification}), and turns a failed check into a \emph{proposal}, a bounded revision that is executed and checked again (Sec.~\ref{sec:repair}). The first rollout that passes all required checks $\mathcal J_{\mathrm{req}}$ conditions $G_\phi$, which runs once with an expanded prompt $\widetilde p$ (Sec.~\ref{sec:control}).

\begin{algorithm}[t]
 \caption{Inference with \method.}
 \label{alg:main}
 \small
 \begin{algorithmic}[1]
  \Require $I_0$, $p$, frozen $G_\phi$, required checks $\mathcal J_{\mathrm{req}}$
  \State Build and validate $\Pi$; compile $\boldsymbol\theta$ (Sec.~\ref{sec:program})
  \For{$k=0,\ldots,3$}
   \State Simulate $\tau^{(k)}$; construct depth control $D_{1:T_g}$
   \State Run checks; collect records $R_j^{(k)}$ (Sec.~\ref{sec:verification})
   \State \textbf{if} $A^{(k)}=1$ \textbf{then return} $G_\phi(I_0,\widetilde p,D_{1:T_g})$
   \State Keep the program if $k=0$ or improved; else roll back
   \State \textbf{if} $k=3$ or a stopping rule holds \textbf{then break}
   \State Replan $\Pi$ or update $\boldsymbol\theta$ by Eq.~\eqref{eq:method_repair}
  \EndFor
  \State \Return fallback $G_\phi(I_0,\widetilde p)$ (Sec.~\ref{sec:control})
 \end{algorithmic}
\end{algorithm}

\subsection{An Editable Physical Program}
\label{sec:program}
The typed physical program $\Pi = (\mathcal{O}, \mathcal{E}, \mathcal{Z}, \mathcal{Y})$ makes explicit which objects take part, in what order events happen, and which relations must hold. Each grounded object in $\mathcal{O}$ has a mask in $I_0$, a physical type (rigid, deformable, granular, or fluid), support and contact relations, and a persistent identity linking it to its simulator state and check records. The event graph $\mathcal E$ has one node per event, naming its agent, patient, expected effect, and checks, with edges that order events by precondition. The required relations $\mathcal Z$ state support, contact, containment, and terminal conditions, and $\mathcal Y$ records the source and validation result of each element. For the falling ball, $\mathcal O$ holds the grabber, ball, sand, and table, and $\mathcal E$ chains release, fall, and collision, where the collision must contact and deform the sand.
Three pretrained models derive $\Pi$ from $I_0$ and $p$. \samthree~\cite{carion2026sam} segments the objects that $p$ names, and \depthanything~\cite{yang2024depth} estimates their relative depth, forming camera-aligned 2.5D geometry~\cite{szeliski2011computer}. Claude Opus 4.8~\cite{anthropic2026claudeopus48}, the VLM used throughout this paper, proposes object roles and the event graph through a schema-constrained interface, and validation rejects missing masks, duplicate identities, conflicting roles, invalid event links, and incompatible physical types before simulation.

Compilation assigns geometry, camera parameters, and execution parameters $\boldsymbol\theta$, such as material properties, initial motion, release timing, and actuation, and maps each object to its Genesis representation. Separating the continuous parameters $\boldsymbol\theta$ from the discrete structure $\Pi$ lets parameter repair change numerical values without changing the event structure. Each tunable parameter $\theta_\ell$ has bounds $[l_\ell,u_\ell]$ and an \emph{owner}, the object or stage it belongs to. After capability checks confirm that Genesis supports the requested interactions and measurements, attempt $k$ executes the specification and yields $\tau^{(k)}$, which records positions, velocities, contacts, applicable transfer or conservation quantities, and whether the solver stayed stable.

\subsection{Hybrid Verification}
\label{sec:verification}

Before synthesis, our hybrid verifier evaluates two criteria about each attempt: does $\tau^{(k)}$ realize the requested event, and does $D_{1:T_g}$ show it to the generator? The VLM judges only event semantics, on rollout frames rendered in the input camera. All other checks measure $\tau^{(k)}$ or $D_{1:T_g}$, and a failed measurement returns a \emph{residual}, the normalized size of its violation, which later scales the repair. The first four check groups below examine $\tau^{(k)}$, and the fifth examines $D_{1:T_g}$.

\paragraph{Event Semantics.}
The VLM checks the identity of the participants, the expected effect, and the order of the events. It scores these aspects separately and keeps the minimum, so a strong score on one aspect cannot mask a failure on another. Semantic failures are localized to object assignments, event dependencies, or release logic.

\paragraph{Physical Process.}
Process checks measure contacts, trajectories, penetration, and applicable transfer or conservation, because a fall may look plausible while an object misses or passes through its support. Failures are localized to object roles, placement, or execution parameters such as initial motion and release timing.

\paragraph{Terminal State.}
Terminal checks measure final support, containment, or settlement, as contact alone does not show that an object comes to rest. Their repairs change the associated geometry and physical parameters while the requirement stays fixed.

\paragraph{Numerical Health.}
Health checks test whether the solver stayed stable, and the measurements above count only for a numerically healthy rollout. Instability routes to solver settings such as the number of substeps.

\paragraph{Control Visibility.}
Because projection and occlusion can hide a correct event from the generator, control checks measure motion direction, event order, the moving regions, and the visible final state on $D_{1:T_g}$. When an editable field is available, their failures go to projection, visibility, or scene construction, and the rebuilt control is checked again.

Each check returns a \emph{check record} that states where the failure lies, what may change, and by how much:
\begin{equation}
 R_j^{(k)}=(s_j,r_j,e_j,\mathcal O_j,[a_j,b_j],\Omega_j),
 \label{eq:method_report}
\end{equation}
where $s_j\in\{\texttt{PASS},\texttt{FAIL}\}$ is the decision of check $j$ at attempt $k$, $r_j$ its residual when measurable, $e_j$ the evidence and criterion, $\mathcal O_j$ the affected objects, $[a_j,b_j]$ the checked interval, and $\Omega_j$ the editable fields.

A check returns \texttt{PASS} only when its required evidence exists and meets the criterion; missing evidence fails without a residual. The acceptance indicator $A^{(k)}$ of attempt $k$ is 1 only if every check in $\mathcal J_{\mathrm{req}}$ passes:
\begin{equation}
 A^{(k)}=
 \begin{cases}
  1, & \text{if } s_j=\texttt{PASS} \text{ for all } j\in\mathcal J_{\mathrm{req}},\\
  0, & \text{otherwise},
 \end{cases}
 \label{eq:method_acceptance}
\end{equation}
so VLM judgments cannot override failed physical measurements, which limits the impact of VLM hallucinations.

\paragraph{Worked Example.}
Fig.~\ref{fig:verification} traces these checks on a \physicsiq sample in which grabbers release a ball and a block onto two pillows, with one column per check group on $\tau^{(k)}$. At $k=1$, the program treats the hanging block as a static fixture, so the VLM, process, and terminal checks fail on the block. The failed release check names the block in $\mathcal O_j$ and its role in $\Omega_j$. Because a role belongs to the program structure, the record calls for replanning rather than a parameter update. With the health failure at $k=0$ and a full pass at $k=2$, $A^{(0)}=A^{(1)}=0$ and $A^{(2)}=1$, so only the third rollout's depth control reaches the generator.

\subsection{Verification-Guided Repair}
\label{sec:repair}

A proposal may change only the fields $\Omega_j$ of a failed check $j$, and the kind of field decides the kind of revision. \emph{Replanning} changes roles, dependencies, or relations in $\Pi$ and reruns compilation and capability checks, as when the fixed block becomes a moving object. \emph{Parameter repair} instead keeps $\Pi$ fixed and changes tunable parameters in $\boldsymbol\theta$, rerunning geometry or control construction when that step failed. Parameter repair scales each edit by its residual:
\begin{equation}
 \theta_\ell'=\operatorname{clip}_{[l_\ell,u_\ell]}
 (\theta_\ell-\alpha_j d_{j\ell}r_j\sigma_\ell),
 \quad \ell\in\Omega_j^\theta,
 \label{eq:method_repair}
\end{equation}
where $\Omega_j^\theta\subseteq\Omega_j$ holds the tunable parameters of failed check $j$, $\alpha_j$ sets the step size, $d_{j\ell}$ is the direction given by a per-check rule, and $\sigma_\ell$ is the parameter scale. A contact check that requires the block-to-pillow distance $z_j$ to stay below a threshold $\eta_j$, for example, has $r_j=\max(0,z_j-\eta_j)/\max(|\eta_j|,\epsilon)$ with a small $\epsilon>0$, so a wider miss yields a larger edit within the bounds. The update is gradient-free, so it needs no differentiable simulator.

Each proposal starts from the \emph{kept program}, initially the first one, and is rechecked against all of $\mathcal J_{\mathrm{req}}$, because an edit can break a requirement that already held. A revision with $A^{(k)}=1$ in Eq.~\eqref{eq:method_acceptance} sends its depth control to the generator. One that fails but improves, by passing more physical checks or as many with a lower largest residual and no other check regressing beyond a tolerance, becomes the kept program; otherwise, we restore the kept program and halve $\alpha_j$. With one initial rollout and at most three revisions, refinement stops on acceptance, after two consecutive rejections, or when no \emph{admissible edit} remains, that is, no tunable field of a failed check can change within its bounds.

\subsection{Depth Control and Generation}
\label{sec:control}
Each rollout is converted into $D_{1:T_g}$ before the control checks run. We remove the moving objects from the initial depth map and reconstruct the static background, so their initial silhouettes do not persist. We then project the simulated objects and particles into the input camera, resolve visibility, and use one depth scale for all frames, so the control changes with scene motion rather than with per-frame rescaling. With the last frame repeated to reach $T_g$ frames, the accepted control conditions the frozen generator:
\begin{equation}
 \widehat V_{1:T_g}=G_\phi(I_0,\widetilde p,D_{1:T_g}),
 \label{eq:generation}
\end{equation}
where the VLM expands $p$ into $\widetilde p$, which keeps the target event and adds visual details of the expected process, such as how objects move and deform, what happens at contact, which objects stay still, and the final state (excerpts in Fig.~\ref{fig:qualitative}). 
When capability checks flag a phenomenon beyond simulator scope, such as pure optics or thermal phase changes, or when no rollout is accepted, \emph{capability routing} sends the input to the \emph{fallback}. The fallback runs $G_\phi(I_0,\widetilde p)$ with the same sampling settings and seed but without depth control, so each input makes one generator call.

\section{Experiments}
\label{sec:experiments}

\physicsiq compares each generated video with a real recording of the same event, whereas \phygen and \vbench score physical plausibility with vision-language evaluators and no reference video.

\begin{table*}[!t]
  \caption{\textbf{Physics-IQ results.} Components and domain scores use the Original protocol. The Verified column gives the overall score on Physics-IQ Verified~\cite{radsch2026verified}. Domain headers give scenario counts, with three views each; Therm.\ and Mag.\ denote thermodynamics and magnetism. Red and blue mark the best and second-best values; ties share a color. Each backbone is followed by \method with that backbone. OmniWeaving uses I2V for the baseline and \texttt{tiv2v} with \method.}
  \label{tab:physicsiq}
  \centering
  \scriptsize
  \setlength{\tabcolsep}{2.5pt}
  \begin{tabular*}{\textwidth}{@{\extracolsep{\fill}}lrrrrrrrrrrr@{}}
    \toprule
    & \multicolumn{4}{c}{Original: components} & \multicolumn{5}{c}{Original: domains $\uparrow$} & \multicolumn{2}{c}{Overall $\uparrow$} \\
    \cmidrule(lr){2-5}\cmidrule(lr){6-10}\cmidrule(lr){11-12}
    Method / backbone & SP $\uparrow$ & ST $\uparrow$ & WS $\uparrow$ & MSE $\downarrow$
      & Solid (38) & Fluid (15) & Optics (8) & Therm.\ (3) & Mag.\ (2) & Original & Verified \\
    \midrule
    \cogvideoxi~\cite{yang2025cogvideox} & 41.11 & 23.36 & 27.21 & 0.48 & 36.57 & 33.08 & 15.00 & 13.12 & 14.07 & 30.07 & 31.58 \\
    \waniiv~\cite{wan2025wan} & 52.75 & 15.32 & 36.18 & 0.55 & 38.70 & 32.11 & 26.86 & 22.47 & 11.39 & 34.20 & 33.57 \\
    PhyT2V~\cite{xue2025phyt2v} & 49.15 & 21.72 & 32.80 & 0.58 & 38.58 & 33.21 & 25.44 & 13.86 & 15.18 & 33.98 & 34.03 \\
    VLIPP~\cite{yang2025vlipp} (CogVideoX) & 49.10 & 24.15 & 32.23 & 0.56 & 42.30 & 34.10 & 16.90 & 8.80 & 13.40 & 34.60 & 34.50 \\
    VLIPP~\cite{yang2025vlipp} (\wanvace) & 50.40 & 22.90 & 33.83 & 0.60 & 41.78 & 30.24 & 26.31 & 13.05 & 13.26 & 35.11 & 34.52 \\
    CoECT~\cite{wang2026chain} (generated frame) & 36.85 & 13.94 & 23.66 & 0.87 & 31.74 & 26.89 & 16.32 & 10.24 & 11.58 & 23.95 & 26.03 \\
    CoECT~\cite{wang2026chain} (observed frame) & 44.53 & 17.36 & 29.71 & 0.61 & 35.61 & 31.28 & 20.15 & 11.87 & 13.42 & 29.92 & 31.07 \\
    SketchVerify~\cite{huang2025planning} & 50.73 & 23.10 & 34.18 & 0.52 & 40.08 & 34.72 & 26.96 & 14.94 & 17.42 & 35.48 & 35.34 \\
    NEWTON~\cite{feng2026newton} & 52.31 & 24.25 & 35.59 & 0.49 & 40.91 & 36.11 & 28.36 & 15.73 & 18.98 & 36.89 & 36.47 \\
    \midrule
    \wanvace~\cite{wan2025wan,jiang2025vace} & 41.44 & 15.07 & 27.67 & 0.73 & 30.42 & 24.23 & 28.07 & 14.34 & 12.56 & 27.33 & 28.39 \\
    \method + \wanvace~\cite{wan2025wan,jiang2025vace} & \second{55.08} & 26.41 & 38.00 & 0.43 & 43.46 & 39.94 & \second{31.27} & 18.22 & 22.65 & 39.40 & 38.32 \\
    \midrule
    OmniWeaving~\cite{pan2026omniweaving} (I2V) & 40.95 & 24.81 & 28.70 & 1.79 & 35.39 & 25.57 & 20.23 & 20.64 & 21.80 & 29.69 & 28.64 \\
    \method + OmniWeaving~\cite{pan2026omniweaving} & 48.61 & 20.76 & 34.07 & 0.77 & 40.05 & 32.20 & 23.02 & 20.92 & 21.80 & 33.71 & 33.21 \\
    \midrule
    Cosmos3-Nano~\cite{agarwal2026cosmos} & 52.71 & \second{31.08} & \second{39.60} & \second{0.37} & \second{44.26} & \second{45.15} & 30.40 & \second{39.76} & \second{29.31} & \second{40.76} & \second{40.92} \\
    \method + Cosmos3-Nano~\cite{agarwal2026cosmos} & \best{55.40} & \best{34.53} & \best{42.10} & \best{0.34} & \best{48.02} & \best{48.31} & \best{31.84} & \best{41.93} & \best{30.57} & \best{43.67} & \best{43.10} \\
    \bottomrule
  \end{tabular*}
\end{table*}

\subsection{Experimental Setup}
\label{sec:benchmarks}

\paragraph{Physics-IQ.}
\physicsiq~\cite{motamed2026physics} records 66 physical events in solid mechanics, fluid dynamics, optics, thermodynamics, and magnetism from three camera views, giving 198 test samples, each recorded twice to capture natural variation. Given only the event prompt and the switch frame, where the observed context ends, \method generates the next five seconds. Spatial, Spatiotemporal, and Weighted Spatial IoU (SP, ST, WS) measure how well the moving regions overlap with the recording, with ST also evaluating timing and WS weighting by motion duration, while RGB mean squared error (MSE) measures appearance. The overall score combines them after normalizing by the agreement between the two recordings, under both the Original protocol~\cite{motamed2026physics} and Physics-IQ Verified~\cite{radsch2026verified}.

\begin{table*}[!t]
  \caption{\textbf{PhyGenBench results.} T2V and I2V denote input protocols. NEWTON~\cite{feng2026newton} uses a trained planner; VideoCoCo~\cite{li2026videococo} uses a LoRA-tuned editor~\cite{hu2021lora}. Parentheses give each method's generator backbone. Red and blue mark the best and second-best values in each column; ties share a color.}
  \label{tab:phygen}
  \centering
  \footnotesize
  \begin{tabular*}{\textwidth}{@{\extracolsep{\fill}}llrrrrr@{}}
    \toprule
    Method & Protocol & Mechanics $\uparrow$ & Optics $\uparrow$ & Thermal $\uparrow$ & Material $\uparrow$ & Overall $\uparrow$ \\
    \midrule
    Pika~\cite{pikalabs2023pika} & T2V & 0.350 & 0.560 & 0.430 & 0.390 & 0.440 \\
    Kling~\cite{kuaishou2024kling} & T2V & 0.450 & 0.580 & 0.500 & 0.400 & 0.490 \\
    \genthree~\cite{runway2024gen3alpha} & T2V & 0.450 & 0.570 & 0.490 & 0.510 & 0.510 \\
    \midrule
    Wan2.1-14B~\cite{wan2025wan} & T2V & 0.360 & 0.530 & 0.360 & 0.330 & 0.400 \\
    CogVideoX-5B~\cite{yang2025cogvideox} & T2V & 0.430 & 0.550 & 0.400 & 0.420 & 0.450 \\
    \ltxvideo-2B~\cite{hacohen2024ltx} & T2V & 0.508 & 0.580 & 0.478 & 0.450 & 0.510 \\
    \wantiiv~\cite{wan2025wan} & T2V & 0.550 & 0.580 & 0.533 & 0.500 & 0.544 \\
    \svdxt~\cite{blattmann2023stable} & I2V & 0.460 & 0.680 & 0.480 & 0.410 & 0.520 \\
    CogVideoX-5B~\cite{yang2025cogvideox} & I2V & 0.480 & 0.690 & 0.430 & 0.410 & 0.520 \\
    \ltxvideo-2B~\cite{hacohen2024ltx} & I2V & 0.470 & 0.650 & 0.460 & 0.370 & 0.500 \\
    \sgiivid~\cite{namekata2025sg} & I2V & 0.520 & 0.690 & 0.510 & 0.390 & 0.540 \\
    \midrule
    PhyT2V~\cite{xue2025phyt2v} (CogVideoX) & T2V & 0.450 & 0.550 & 0.430 & 0.530 & 0.500 \\
    VLIPP~\cite{yang2025vlipp} (\gowithflow~\cite{burgert2025go}) & I2V & 0.550 & 0.710 & 0.600 & 0.530 & 0.600 \\
    NEWTON~\cite{feng2026newton} (\ltxvideo) & T2V & 0.500 & 0.647 & 0.522 & 0.542 & 0.560 \\
    VideoCoCo~\cite{li2026videococo} (OmniWeaving, LoRA) & T2V & 0.558 & 0.613 & 0.511 & 0.525 & 0.558 \\
    \midrule
    \method{} (\wanvace) & I2V & 0.568 & 0.724 & 0.619 & 0.560 & 0.624 \\
    \midrule
    Cosmos3-Nano~\cite{agarwal2026cosmos} & I2V & \second{0.633} & \second{0.740} & \second{0.678} & \second{0.617} & \second{0.671} \\
    \method{} (Cosmos3-Nano) & I2V & \best{0.667} & \best{0.753} & \best{0.722} & \best{0.667} & \best{0.704} \\
    \bottomrule
  \end{tabular*}
\end{table*}

\paragraph{PhyGenBench and VBench-2.0.}
\phygen~\cite{meng2024towards} has 160 prompts covering 27 physical laws in four categories, and its evaluator, PhyGenEval, combines whether the expected phenomenon occurs, whether events follow the right order, and whether the video looks physically natural into a score in $[0,1]$. Each method generates one video per prompt, with a common input image for image-conditioned methods. For \vbench~\cite{zheng2025vbench}, we follow VideoCoCo~\cite{li2026videococo} and report the three physics categories, Mechanics, Thermotics, and Material, which the official evaluator scores through visual question answering.

\paragraph{Implementation Details.}
We pair \method with three frozen generators that take depth control through different interfaces. \wanvace~\cite{wan2025wan,jiang2025vace} is the default, and its baseline receives $I_0$ and $\widetilde p$ without depth control. Cosmos3-Nano~\cite{agarwal2026cosmos} uses its built-in depth conditioning without additional LoRA weights~\cite{hu2021lora}, and OmniWeaving~\cite{pan2026omniweaving} takes depth as a video input in its \texttt{tiv2v} mode. Please refer to \suppref{sec:supp_implementation} for the full implementation details.

\begin{figure*}[!t]
 \centering
 \includegraphics[width=\textwidth]{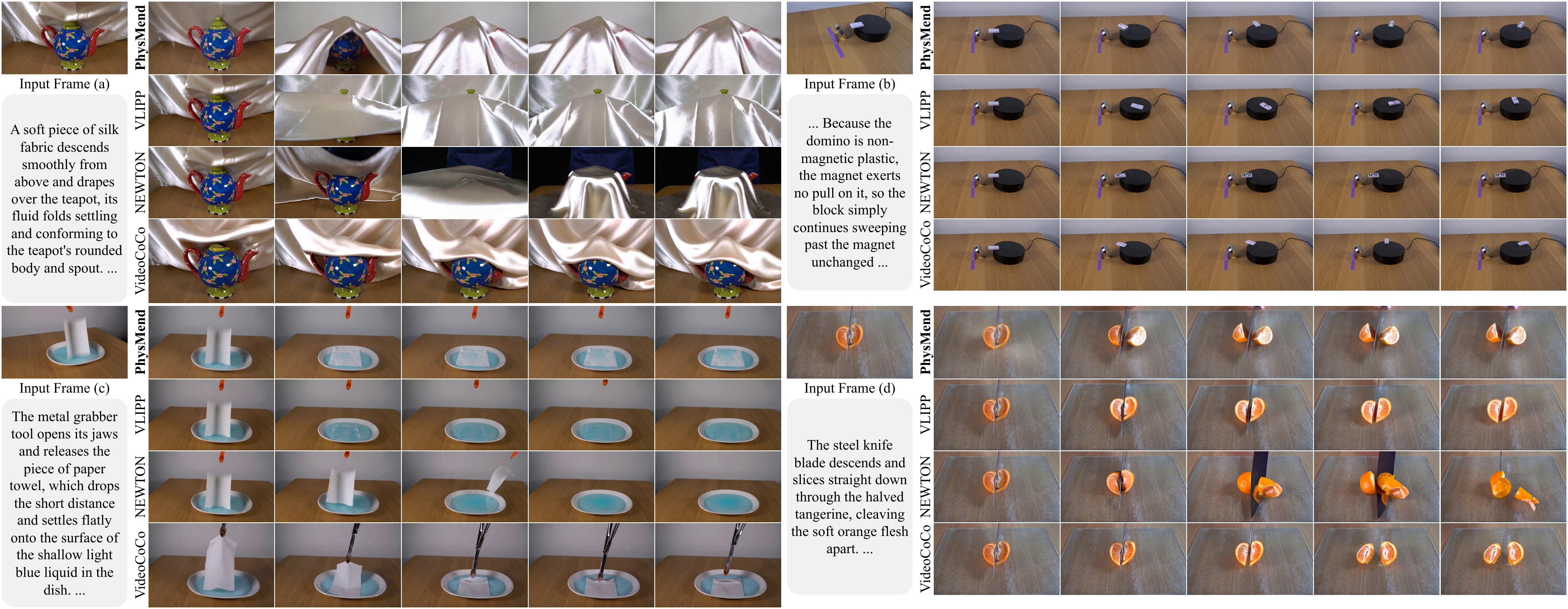}
 \caption{\textbf{Qualitative comparisons on Physics-IQ.} We compare \method with VLIPP~\cite{yang2025vlipp}, NEWTON~\cite{feng2026newton}, and VideoCoCo~\cite{li2026videococo} on four cases. The cases are (a) silk covering a teapot, (b) a plastic domino rotating past a magnet, (c) a paper towel released onto liquid, and (d) a knife cutting a tangerine. Each panel shows the input image, an expanded-prompt excerpt, and five chronological frames per method. Ellipses indicate omitted prompt text. \method occupies the first row. Frames keep their original aspect ratios.}
 \label{fig:qualitative}
\end{figure*}

\subsection{Main Results}
\label{sec:physicsiq_results}

\paragraph{Physics-IQ.}
\method improves all three generators under both the Original and Verified protocols (Tab.~\ref{tab:physicsiq}). On the Original overall score, the relative gain is largest for \wanvace, 44\% (from 27.33 to 39.40): alone it scores below most compared methods, but with \method it outperforms every prior physics-aware method in the table, including VLIPP on the same backbone (35.11) and NEWTON (36.89). OmniWeaving gains 14\% (from 29.69 to 33.71), and Cosmos3-Nano, the strongest baseline, gains 7\% (from 40.76 to 43.67). With Cosmos3-Nano, \method achieves the best value in every column, and with both \wanvace and Cosmos3-Nano, every domain improves.

\paragraph{PhyGenBench.}
\label{sec:phygen_results}
\method improves Cosmos3-Nano by 5\% relative (from 0.671 to 0.704), with gains in every category, most in Material and Thermal, and achieves the highest overall score among the methods in Tab.~\ref{tab:phygen}. With \wanvace, it outperforms PhyT2V, VLIPP, and the agentic workflows \mbox{NEWTON} and VideoCoCo in every category, and exceeds the latter two by more than 11\% relative overall (0.624 vs.\ 0.560 and 0.558).

\paragraph{VBench-2.0.}
\label{sec:vbench_results}
Without fine-tuning the generator, \method raises the OmniWeaving average by 46\% relative (from 52.18 to 76.16), improves all three categories, and nearly doubles the Material score (from 41.67 to 81.54; Tab.~\ref{tab:vbench}). Its average is comparable to those of VideoCoCo (77.88), which uses a LoRA-tuned editor, and CogVideoX-1.5 (77.04), while its Material score exceeds VideoCoCo's by more than 13 points (81.54 vs.\ 68.42). \method reaches this level under a fixed budget of one generator call per input.

\paragraph{Analysis.}
Because the \wanvace baseline already receives the expanded prompt $\widetilde p$, its gain comes from the accepted depth control. Among the \physicsiq components, its Spatiotemporal IoU rises the most, by 75\% relative (from 15.07 to 26.41), so the generated motion overlaps the recording more closely in both space and time. Its largest absolute domain gains occur in fluids and solids, the two domains where physical measurements outperform VLM judgments in the verification ablation (Sec.~\ref{sec:ablations}).

\begin{table}[!htbp]
  \caption{\textbf{VBench-2.0 physics results.} Mech., Therm., and Mat. denote Mechanics, Thermotics, and Material. Scores are percentages; AVG is their unweighted mean. Higher is better. Red and blue mark the best and second-best scores.}
  \label{tab:vbench}
  \centering
  \footnotesize
  \begin{tabular*}{\columnwidth}{@{\extracolsep{\fill}}lrrrr@{}}
    \toprule
    Model & Mech. $\uparrow$ & Therm. $\uparrow$ & Mat. $\uparrow$ & AVG $\uparrow$ \\
    \midrule
    Sora~\cite{brooks2024video} & 62.22 & 43.36 & 64.94 & 56.84 \\
    Kling 1.6~\cite{kuaishou2024kling} & 65.55 & 59.46 & 68.00 & 64.34 \\
    HunyuanVideo~\cite{kong2024hunyuanvideo} & 76.09 & 56.52 & 64.37 & 65.66 \\
    CogVideoX-1.5~\cite{yang2025cogvideox} & 80.80 & \second{67.13} & \best{83.19} & \second{77.04} \\
    OmniWeaving~\cite{pan2026omniweaving} & 62.79 & 52.08 & 41.67 & 52.18 \\
    \quad + VideoCoCo~\cite{li2026videococo} & \best{92.31} & \best{72.92} & 68.42 & \best{77.88} \\
    \midrule
    OmniWeaving + \method & \second{81.41} & 65.54 & \second{81.54} & 76.16 \\
    \bottomrule
  \end{tabular*}
\end{table}

\subsection{Ablation Studies}
\label{sec:ablations}

All ablations use \physicsiq and \wanvace with a fixed rollout budget (Tab.~\ref{tab:ablations}).

\begin{table}[!htbp]
  \caption{\textbf{Ablations on Physics-IQ.} Original scores with \wanvace~\cite{wan2025wan,jiang2025vace} and a fixed rollout budget. The first row is \wanvace alone, with the expanded prompt $\widetilde p$ and no depth control. The first three failure-handling variants return simulated depth even if no rollout passes. Red and blue mark the best and second-best values per group.}
  \label{tab:ablations}
  \centering
  \footnotesize
  \begin{tabular*}{\columnwidth}{@{\extracolsep{\fill}}lr@{}}
    \toprule
    Variant & Score $\uparrow$ \\
    \midrule
    \wanvace alone & 27.33 \\
    \midrule
    \multicolumn{2}{@{}l}{\textit{System components}} \\
    Untyped simulation & 31.20 \\
    Typed simulation & 34.26 \\
    Verified simulation & \second{37.21} \\
    Full \method & \best{39.40} \\
    \midrule
    \multicolumn{2}{@{}l}{\textit{Verification evidence}} \\
    VLM only & 36.00 \\
    Measurements only & \second{37.28} \\
    Hybrid verifier & \best{39.40} \\
    \midrule
    \multicolumn{2}{@{}l}{\textit{Failure handling}} \\
    No repair & 37.21 \\
    Global rebuild & 37.78 \\
    Localized repair & \second{38.66} \\
    Full routing & \best{39.40} \\
    \bottomrule
  \end{tabular*}
\end{table}

\paragraph{System Components.}
Simulated depth without any checks (\emph{Untyped simulation}) already improves \wanvace from 27.33 to 31.20, accounting for about one third of the total gain. Enforcing schema-validated typed programs (\emph{Typed simulation}, 34.26), rollout verification (\emph{Verified simulation}, 37.21), and closed-loop repair with capability routing (\emph{Full \method}, 39.40) each raise the score further in every domain, contributing the remaining two thirds for a 26\% relative gain over untyped simulation (from 31.20 to 39.40).

\paragraph{Verification Evidence.}
\label{sec:ablation_evidence}
With the same generator, routing, and rollout budget, the hybrid verifier scores highest in every domain and overall, about 6\% above measurements alone and 9\% above VLM judgments alone (39.40 vs.\ 37.28 and 36.00). The two sources complement each other across domains: physical measurements score 4\% and 7\% higher than VLM judgments in solids and fluids, where contact, trajectories, and resting states are measured directly, whereas VLM judgments score 4\% to 7\% higher in optics, thermodynamics, and magnetism, whose visual effects are not fully captured by geometry and motion measurements alone.

\paragraph{Repair and Capability Routing.}
\label{sec:ablation_repair}
With the same samples, rollout budget, and maximum number of repair attempts, editing only the fields of the failed check (\emph{Localized repair}, 38.66) outperforms rebuilding the whole program (\emph{Global rebuild}, 37.78) in every domain, and global rebuild improves only slightly on no repair (37.21; Tab.~\ref{tab:ablations}). Unlike these variants, which always return simulated depth, \emph{Full routing} (39.40) sends inputs that fail capability checks or are not accepted within the budget to the fallback, adding further gains led by magnetism (from 20.84 to 22.65) and thermodynamics (from 17.24 to 18.22). Together, localized repair and capability routing raise the no-repair score by 6\% relative (from 37.21 to 39.40).

\subsection{Qualitative Results}
\label{sec:qualitative}

Fig.~\ref{fig:qualitative} compares four physical events. With \method, the silk settles over the teapot (a), the plastic domino follows the platform's rotation without being pulled by the magnet (b), the released towel comes to rest on the liquid (c), and the knife separates the tangerine while the scene stays consistent (d). VideoCoCo produces a comparable cut in (d) but leaves much of the teapot exposed and keeps the towel in the gripper, and NEWTON moves the towel away from the dish.

\section{Conclusion}
\label{sec:conclusion}
We presented \method, a training-free framework that verifies and locally repairs typed physical programs before a single pass through a frozen video generator. Its hybrid verifier combines VLM-based event judgments with direct physical and depth-visibility checks to guide residual-scaled updates. Experiments on \physicsiq, \phygen, and \vbench demonstrate consistent gains across video generative models with different architectures. Among the compared methods, \method achieves state-of-the-art results on \physicsiq and \phygen and remains comparable to state-of-the-art methods on \vbench under a fixed generation budget.

\label{page:main-end}

\section*{LLM Usage Statement}
We used Claude Opus 5.5 (Anthropic) to polish the author-written text throughout the paper and to help edit and lay out all figures, including writing plotting code. Claude Opus 5.5 generated no technical content, claims, citations, or data, and it was not used for research ideation, method design, experiments, or result analysis. In the research itself, our method uses Claude Opus 4.8 as its VLM: it proposes object roles and the event graph (Sec.~\ref{sec:program}), serves as the semantic judge during verification (Sec.~\ref{sec:verification}), and expands the input prompt (Sec.~\ref{sec:control}). The authors verified all LLM-assisted outputs and take full responsibility for this paper.

{
  \small
  \setlength{\bibsep}{0pt}
  \ifdefined\fullbblfile
  \InputIfFileExists{\fullbblfile}{}{%
    \InputIfFileExists{main.bbl}{}{%
      \PackageError{PhysMend}{Missing \fullbblfile}%
        {Build main.tex into output/cvpr first, e.g. with scripts/build_cvpr_split.sh.}}}
  \else
    \bibliographystyle{ieeenat_fullname}
    \bibliography{references}
  \fi
}
\clearpage
\fi

\ifdefined\mainpaperonly
\else
\appendix
\ifdefined\standalonesupplement
  \continuemaincounters
\else
  \recordmaincounters
\fi
  \twocolumn[{%
    \renewcommand\twocolumn[1][]{#1}%
    {\maketitlesupplementary\par}%
    \section*{Contents}
    The table below lists each section of this supplement, what it contains, the part of the main paper that it supports, and the page where it starts.\par
    \vspace{6pt}
    {\small
    \begin{tabularx}{\textwidth}{@{}l>{\raggedright\arraybackslash}p{4.3cm}>{\raggedright\arraybackslash}Xlc@{}}
      \toprule
      Sec. & Title & What it contains & Main paper & Page \\
      \midrule
      \ref{sec:supp_program} & \supptitle{sec:supp_program} & Program fields and grounding, the recorded ball-in-sand program, compilation and capability checks, and the exported rollout & Sec.~\mainref{sec:program} & \pageref{sec:supp_program} \\
      \ref{sec:supp_verification} & \supptitle{sec:supp_verification} & Check records and acceptance, semantic and physical predicates, parameter updates, the correction loop and stopping rules, and the generator interface & Secs.~\mainref{sec:verification}--\mainref{sec:control} & \pageref{sec:supp_verification} \\
      \ref{sec:supp_metrics} & \supptitle{sec:supp_metrics} & Protocols, metric definitions, and score aggregation of \physicsiq, \phygen, and \vbench & Sec.~\mainref{sec:benchmarks} & \pageref{sec:supp_metrics} \\
      \ref{sec:supp_omniweaving} & \supptitle{sec:supp_omniweaving} & The OmniWeaving interface used with \method & Sec.~\mainref{sec:benchmarks}, Tab.~\mainref{tab:physicsiq} & \pageref{sec:supp_omniweaving} \\
      \ref{sec:supp_ablations} & \supptitle{sec:supp_ablations} & Domain scores for system components, verification evidence, repair and capability routing, and control representation & Sec.~\mainref{sec:ablations} & \pageref{sec:supp_ablations} \\
      \ref{sec:supp_efficiency} & \supptitle{sec:supp_efficiency} & The simulation budget per input, and generation calls, time, and Physics-IQ score compared with NEWTON & Secs.~\mainref{sec:repair} and \mainref{sec:control}, Tab.~\mainref{tab:physicsiq} & \pageref{sec:supp_efficiency} \\
      \ref{sec:supp_seedance} & \supptitle{sec:supp_seedance} & Seedance 2.5 guided by particle-motion, coarse-RGB, or depth videos on ten \physicsiq scenes & Sec.~\mainref{sec:control} & \pageref{sec:supp_seedance} \\
      \ref{sec:supp_reporting} & \supptitle{sec:supp_reporting} & Generators, baselines, generation settings, and reproducibility & Sec.~\mainref{sec:benchmarks} & \pageref{sec:supp_reporting} \\
      \ref{sec:supp_figures} & \supptitle{sec:supp_figures} & The examples shown in the overview and verification figures & Figs.~\mainref{fig:overview} and \mainref{fig:verification} & \pageref{sec:supp_figures} \\
      \ref{sec:supp_qualitative} & \supptitle{sec:supp_qualitative} & Eight more \physicsiq comparisons with VLIPP, NEWTON, and VideoCoCo & Sec.~\mainref{sec:qualitative}, Fig.~\mainref{fig:qualitative} & \pageref{sec:supp_qualitative} \\
      \ref{sec:limitations} & \supptitle{sec:limitations} & Limits of the method & Secs.~\mainref{sec:method} and \mainref{sec:experiments} & \pageref{sec:limitations} \\
      \bottomrule
    \end{tabularx}\par}
    \vspace{12pt}
  }]
  \ifdefined\fillzeros
    \xdef\supplinenumberoffset{\the\csname c@pagewiselinenumber\endcsname}%
    \renewcommand\thelinenumber{%
      \fillzeros[3]{\numexpr\value{linenumber}-\supplinenumberoffset\relax}}%
  \fi
\section{Program and Simulation Details}
\label{sec:supp_program}

\subsection{Program Representation and Grounding}

The physical program is
\begin{equation}
 \Pi=(\mathcal O,\mathcal E,\mathcal Z,\mathcal Y),
 \label{eq:program}
\end{equation}
where $\mathcal O$ is the grounded object set, $\mathcal E$ the event graph, $\mathcal Z$ the required support, contact, containment, and terminal relations, and $\mathcal Y$ the record of each element's source and validation result. Each event specifies its participants, preconditions, dependencies, expected effect, physical primitives, and checks.

The inputs are an initial image $I_0$ and a text prompt $p$ that describes the physical event to generate. Grounding links this description to the scene: each entity that $p$ names receives a persistent identity, a visible mask, a confidence value, relative depth, and support or containment cues. Every object record stores
\begin{quote}
\small
\texttt{id, mask identifier, class name, semantic role, physical type, material category, state phases, initial motion, support relations, contact relations, and provenance.}
\end{quote}
Every event record stores
\begin{quote}
\small
\texttt{event id, event type, agent, patient, physical primitives, preconditions, causal dependencies, expected effect, and required status.}
\end{quote}

Grounding combines object masks from \samthree~\cite{carion2026sam}, relative depth from \depthanything~\cite{yang2024depth}, and roles and event structure proposed by Claude Opus 4.8~\cite{anthropic2026claudeopus48}, the vision-language model (VLM) used throughout this paper. Together with visible relations, these estimates define camera-aligned 2.5D geometry~\cite{szeliski2011computer}. Validation rejects missing masks, duplicate identities, role conflicts, invalid event links, incompatible physical types, and primitives absent from the capability registry. VLM output never reaches the simulator without this validation.

Tab.~\ref{tab:supp_program_example} lists the recorded program of the ball-in-sand example from Sec.~\mainfloatref{sec:program}\inmainpaper. Its four objects and three events match that description; masks, provenance, and validation records are omitted.

\begin{table*}[t]
  \caption{Recorded program of the ball-in-sand example. Identifiers and required checks are listed as recorded.}
  \label{tab:supp_program_example}
  \centering
  \footnotesize
  \begin{tabular}{@{}llll@{}}
    \toprule
    Object & Class & Physical type & Support and contact relations \\
    \midrule
    \texttt{ball} & tennis ball & rigid & supported by and held by \texttt{grabber} \\
    \texttt{kinetic\_sand} & kinetic sand pile & granular & supported by and rests on \texttt{table} \\
    \texttt{grabber} & grabber tool & rigid & holds \texttt{ball} \\
    \texttt{table} & wooden table & boundary & supports \texttt{kinetic\_sand} \\
    \bottomrule
  \end{tabular}

  \vspace{6pt}
  \begin{tabular}{@{}llllp{4.4cm}p{3.6cm}@{}}
    \toprule
    Event & Agent & Patient & Precondition & Expected effect & Required checks \\
    \midrule
    \texttt{event\_1} (release) & \texttt{grabber} & \texttt{ball} & none & \raggedright grabber opens and lets go of the ball, ending the support relation & \raggedright \texttt{SUPPORT}, \texttt{CAUSAL\_ORDER} \tabularnewline
    \texttt{event\_2} (gravity fall) & \texttt{ball} & none & \texttt{event\_1} & \raggedright ball accelerates downward toward the sand pile & \raggedright \texttt{KINEMATIC\_FREE}, \texttt{CAUSAL\_ORDER} \tabularnewline
    \texttt{event\_3} (collision) & \texttt{ball} & \texttt{kinetic\_sand} & \texttt{event\_2} & \raggedright ball strikes the sand pile, deforming its surface and possibly settling into it & \raggedright \texttt{CONTACT}, \texttt{DEFORM}, \texttt{CAUSAL\_ORDER} \tabularnewline
    \bottomrule
  \end{tabular}
\end{table*}

\subsection{Compilation and Capability Checks}

Compilation produces the simulation specification
\begin{equation}
 \Theta=(\Pi,\mathcal M,D_0,K,\boldsymbol\theta,B),
 \label{eq:compile}
\end{equation}
where $\mathcal M$ contains object masks, $D_0$ is the initial depth, $K$ the assumed camera parameters, $\boldsymbol\theta$ the execution parameters, and $B$ the backend representation. Each scalar $\theta_\ell$ has bounds $[l_\ell,u_\ell]$, a source, an owner, and a flag that marks whether repair may change it; flagged scalars are the tunable parameters.

A deterministic capability registry determines whether the selected representations support the requested interactions, units, parameter ranges, and measurements. Tab.~\ref{tab:supp_mapping} lists the Genesis~\cite{Genesis} representation of each physical type. Inputs that fail validation or capability checks go to the fallback.

\begin{table}[!htbp]
  \caption{Physical type to Genesis~\cite{Genesis} representation.}
  \label{tab:supp_mapping}
  \centering
  \small
  \begin{tabular}{@{}ll@{}}
    \toprule
    Physical type & Representation \\
    \midrule
    Rigid or brittle object & Rigid primitive \\
    Deformable object & Elastic MPM material~\cite{sulsky1994particle} \\
    Granular material & MPM sand \\
    Fluid & SPH liquid~\cite{gingold1977smoothed} \\
    Boundary (support or container) & Fixed rigid geometry \\
    \bottomrule
  \end{tabular}
\end{table}

MPM and SPH denote the Material Point Method~\cite{sulsky1994particle} and Smoothed Particle Hydrodynamics~\cite{gingold1977smoothed}, respectively. The compiler records the material and execution parameters needed by the selected representation, including mass, density, friction, damping, elasticity, viscosity, initial velocity, release timing, coupling values, solver resolution, and substeps. Recording the owner and bounds of each tunable parameter restricts later corrections to these declared fields.

\subsection{Execution and Exported Evidence}

Each simulation attempt runs in a fresh process that lifts masks into camera-aligned geometry and builds the selected physical representations. It sets the hidden depth extent from scene relations, normalizes the scene into a stable solver domain, and checks initialization before applying event controls and exporting 80 frames. Alongside this state sequence, the simulator exports positions, depths, visible states, contacts, contact forces, support and containment relations, particle ledgers, initialization diagnostics, and numerical health for verification.

For attempt $k\in\{0,\ldots,3\}$, the rollout is
\begin{equation}
 \tau^{(k)}=\{\mathbf x_{i,t},\mathbf v_{i,t},\mathcal C_t,
                  \mathcal L_t,\mathcal H_t\}_{t=1}^{T_s},
 \label{eq:trajectory}
\end{equation}
where $i$ indexes objects or particles, $\mathcal C_t$ records contacts and relations, $\mathcal L_t$ records applicable conservation or transfer ledgers, and $\mathcal H_t$ records numerical health. The $T_s$ exported frames are distinct from internal integration steps and the $T_g$ frames produced by the generator.

\section{Verification and Repair Details}
\label{sec:supp_verification}

\subsection{Check Records and Acceptance}

Each configured check returns a decision, its evidence, the affected objects, a time interval, the editable fields, and a residual when measurable:
\begin{equation}
 R_j^{(k)}=(s_j,r_j,e_j,\mathcal O_j,[a_j,b_j],\Omega_j).
 \label{eq:report}
\end{equation}
Here $j$ indexes the check and $k$ the simulation attempt; we suppress $k$ on the record fields for readability. The decision $s_j$ is either \texttt{PASS} or \texttt{FAIL}, $r_j$ is the normalized residual when measurable, and $e_j$ records the evidence and criterion. The remaining fields identify the affected objects $\mathcal O_j$, checked interval $[a_j,b_j]$, and editable fields $\Omega_j$. A check passes only if its required evidence exists and meets the criterion; otherwise it fails, and $e_j$ stores the reason. A measurable failure stores the signed error and its threshold, whereas a failure for missing evidence has no numerical residual. Physical measurements count only for a numerically healthy simulation.

For the required checks $\mathcal J_{\mathrm{req}}$, the acceptance indicator of attempt $k$ is
\begin{equation}
 A^{(k)}=
 \begin{cases}
  1, & \text{if } s_j=\texttt{PASS} \text{ for all } j\in\mathcal J_{\mathrm{req}},\\
  0, & \text{otherwise}.
 \end{cases}
 \label{eq:accepted}
\end{equation}
Acceptance applies to the rollout and its depth control.

\subsection{Semantic and Physical Predicates}

Process checks cover contacts, trajectories, response timing, penetration, continuity, and applicable transfer or conservation. Terminal checks evaluate the requested support, containment, or settling state. The following predicates illustrate the semantic and physical measurements. Semantic scores come from the VLM.

For each event, the semantic score is
\begin{equation}
\psi^{\mathrm{sem}}=
\min\left(\psi^{\mathrm{id}},\psi^{\mathrm{effect}},\psi^{\mathrm{order}}\right).
\end{equation}
Here the three component scores assess participant identity, expected effect, and event order. Taking their minimum keeps a strong score on one aspect from masking a failure on another. A semantic check passes when $\psi^{\mathrm{sem}}$ is at least 0.75, and a semantic judgment cannot override a failed physical check.

For the projected point sets $X_t$ and $X'_t$ of two objects at frame $t$, the contact distance is
\begin{equation}
\delta(t)=\min_{\mathbf y\in X_t,\mathbf y'\in X'_t}
\lVert \mathbf y-\mathbf y'\rVert_2.
\end{equation}
The contact predicate passes when Genesis reports contact or the minimum projected distance is within the contact threshold. These branches provide different evidence: solver contact concerns simulated geometry, whereas projected distance measures image-space proximity.

For fluid transfer with total particle count $N$, the transfer, leakage, and retention ratios are
\begin{equation}
\rho_{\mathrm{transfer}}(t)=\frac{N_{\mathrm{receiver}}(t)}{N},
\end{equation}
\begin{equation}
\rho_{\mathrm{leak}}(t)=
\frac{N_{\mathrm{leak}}(t)}
{\max(N_{\mathrm{released}}(t),1)},
\end{equation}
\begin{equation}
\rho_{\mathrm{retain}}=
\frac{N_{\mathrm{receiver}}(T_s)}
{\max_t N_{\mathrm{receiver}}(t)}.
\end{equation}
Here $N_{\mathrm{receiver}}$, $N_{\mathrm{released}}$, and $N_{\mathrm{leak}}$ count receiver, released, and leaked particles; $T_s$ is the final exported simulation frame. Transfer and leakage have different denominators and therefore need not sum to one. Retention is defined only when $\max_t N_{\mathrm{receiver}}(t)>0$; if no material enters, it supplies no numerical residual. Settling requires low terminal speed together with the requested support or containment relation.

\subsection{Failure Localization and Parameter Updates}
\label{sec:supp_repair}

\begin{figure*}[!t]
  \centering
  \includegraphics[width=\textwidth]{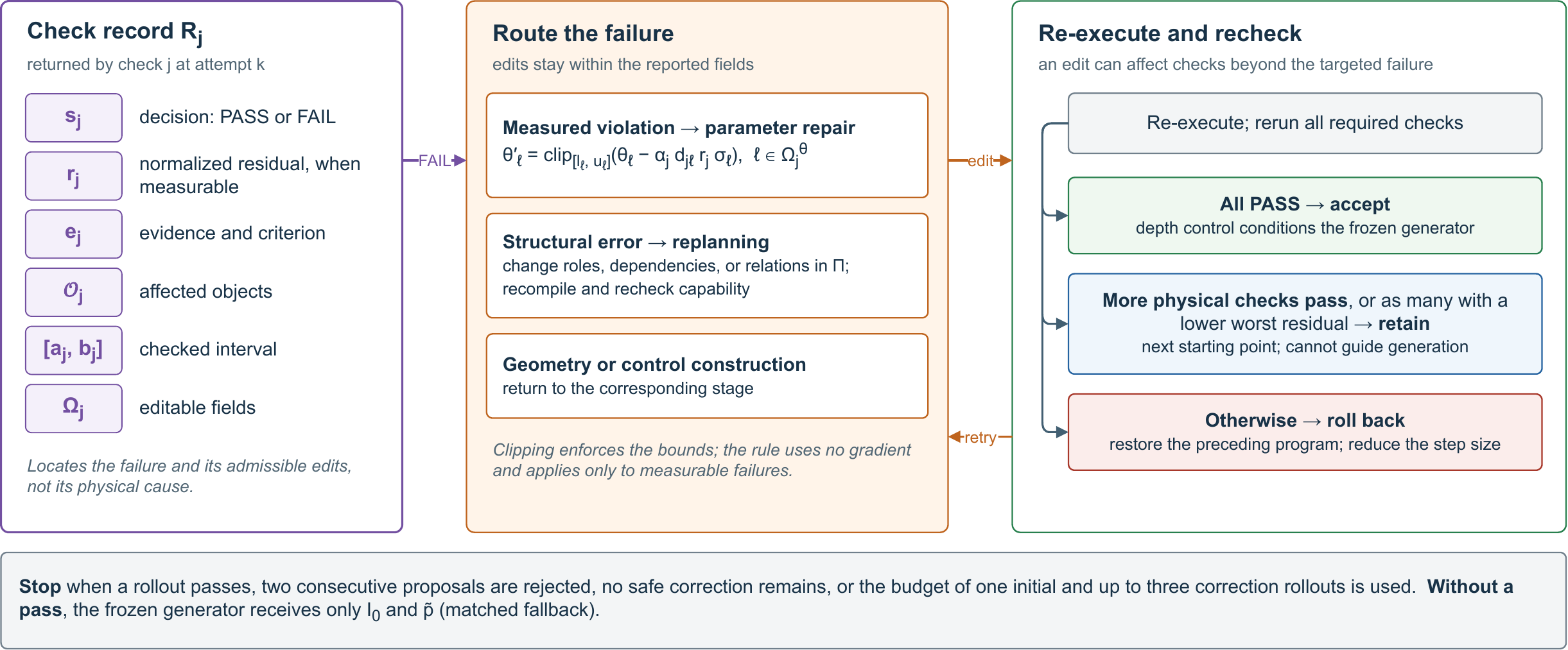}
  \caption{\textbf{From check record to bounded repair.} A failed check reports its decision, residual, evidence, affected objects, checked interval, and editable fields. Measured violations scale bounded edits of the reported parameters, structural errors trigger replanning, and construction failures return to their stage. Every revision is re-executed against all required checks and is accepted, retained as a new starting point, or rolled back.}
  \label{fig:supp_repair}
\end{figure*}

Failure localization uses recorded field ownership to associate a failed check with program, compilation, initialization, numerical, actuation, or control fields. This association guides a \emph{proposal}, meaning a revision to the program structure or its execution parameters. Fig.~\ref{fig:supp_repair} traces this path from a check record to a re-executed revision. Replanning changes discrete structure and therefore repeats compilation and capability checks, whereas parameter repair keeps $\Pi$ fixed and modifies only declared tunable parameters associated with the failure. Correction stops when no admissible edit remains.

For an upper-bound check $z_j\leq\eta_j$, where $z_j$ is the measured quantity and $\eta_j$ its threshold, the normalized violation is
\begin{equation}
 r_j=\frac{\max(0,z_j-\eta_j)}{\max(|\eta_j|,\epsilon)},
 \qquad \epsilon>0.
 \label{eq:repair_residual}
\end{equation}
Here $\epsilon$ is a positive denominator guard. A lower-bound check $z_j\geq\eta_j$ uses $\max(0,\eta_j-z_j)$ in the numerator, so a satisfied bound has zero residual and a violated bound has a positive one. The direction $d_{j\ell}$ in the update below determines which way to change a parameter; it is separate from the residual magnitude. This normalization requires the measurement to exist.

For a tunable parameter $\theta_\ell\in[l_\ell,u_\ell]$, the update is
\begin{equation}
 \theta_\ell'=\operatorname{clip}_{[l_\ell,u_\ell]}
 (\theta_\ell-\alpha_j d_{j\ell}r_j\sigma_\ell),\quad \ell\in\Omega_j^\theta,
 \label{eq:repair}
\end{equation}
where $j$ indexes a failed check with a measurable residual, $\Omega_j^\theta\subseteq\Omega_j$ contains its tunable parameters, $r_j$ is its normalized residual, $\alpha_j$ the step factor, $d_{j\ell}$ the update direction given by a per-check rule, and $\sigma_\ell$ the scale of the parameter. Values outside the selected set remain unchanged. Normalization sets the size of the violation, whereas the choice of editable parameters, directions, and scales is specific to each check. The rule bounds each proposal and uses no gradient. A failure caused by missing evidence has no residual and cannot use this update; if no admissible edit remains, the input goes to the fallback.

A proposal is kept when it passes more physical checks, or passes as many with a lower largest residual while no other check regresses beyond a tolerance; process and terminal residuals take priority. Otherwise the kept program is restored and the step factor is halved, down to a floor of 0.05. When numerical health fails, the proposal multiplies the solver substeps by 1.5, rounded up and capped at 1024. Re-execution tests each of these updates.

A kept revision becomes the starting point for the next proposal. Only a rollout that passes every required check sends its depth control to the generator, and the loop stops at the first such rollout.

\subsection{Correction Loop and Stopping Rules}
\label{sec:supp_loop}

\begin{algorithm}[t]
 \caption{Inference procedure of \method}
 \label{alg:physmend}
 \small
 \begin{algorithmic}[1]
  \Require $I_0,p$; frozen $G_\phi$; checks and field ownership
  \State Expand $p$ into the generator prompt $\widetilde p$
  \State Ground and compile $(I_0,p)$ into $\Theta$
  \If{validation or capability checks fail}
    \State \Return image-and-prompt fallback
  \EndIf
  \State Set the current program to $\Theta$; rejection count $c\gets0$
  \For{$k=0,\ldots,3$}
    \State Execute the current program; construct control; run all checks
    \State Set each decision to \texttt{PASS} or \texttt{FAIL} (Eq.~\eqref{eq:report})
    \If{every required check returns \texttt{PASS}}
      \State \Return $G_\phi(I_0,\widetilde p,D_{1:T_g})$
    \EndIf
    \If{$k>0$}
      \If{the proposal improves on the kept program}
        \State Keep the current program and its check records; $c\gets0$
      \Else
        \State Restore the kept program and its check records; halve the step factor; $c\gets c+1$
      \EndIf
    \Else
      \State Keep the initial program and its check records
    \EndIf
    \If{$k=3$ or $c=2$ or no admissible edit remains}
      \State \textbf{break}
    \EndIf
    \State Propose a revision of the kept program and make it current
    \State Recheck compilation and capability; \textbf{break} if invalid
  \EndFor
  \State \Return image-and-prompt fallback
 \end{algorithmic}
\end{algorithm}

Alg.~\ref{alg:physmend} separates acceptance, which sends the depth control to the generator, from keeping, which only makes a revision the next starting point. The initial program is the first kept program, and a rejected proposal restores both the kept program and its check records. A structural revision passes compilation and capability checks again before it runs. RGB generation stays outside the loop, so each input makes one generator call.

\subsection{Depth Control and Generator Interface}
\label{sec:supp_control}

The renderer removes moving pixels from the initial depth map, reconstructs the static background depth, rasterizes objects and particles into the camera view, and resolves visibility. Splatting points together with their object labels, normalizing depth over the whole clip, and Gaussian smoothing~\cite{szeliski2011computer} with a one-pixel standard deviation then produce the dense control sequence. Control checks reject wrong direction, reversed order, missing events, empty frames, absent dynamic foreground, severe flicker, static control for a dynamic target, and an invalid final visible state.

Control is checked separately from simulation because a rollout can pass the particle ledger while the camera view does not show the outcome. In a pouring case, 96.11\% of the particles reach the receiver, yet the receiver appears only shallowly filled in the camera view.

The default generator is frozen Wan-based VACE (\wanvace)~\cite{wan2025wan,jiang2025vace}. For it, the $T_s=80$ exported simulation frames at 16 fps become $T_g=81$ generator frames by repeating the last depth frame. The generator receives the initial image, the expanded prompt, and the accepted depth control, and its weights $\phi$ stay frozen. \method is training-free: it trains no model and uses the upstream models and conditioning adapters as released. The fallback uses the same generator, sampling settings, and seed with the image and expanded prompt but no depth, so every input receives one final RGB output.

At inference, \method uses only the event prompt $p$ and the initial image $I_0$ from the benchmark; its checks and parameter updates rely on the estimated scene and the simulated rollout.

\section{Benchmarks and Evaluation Metrics}
\label{sec:supp_metrics}

\subsection{Benchmark Protocols}
\label{sec:supp_benchmarks}

\paragraph{Physics-IQ.}
\physicsiq~\cite{motamed2026physics} contains 66 scenarios across solid mechanics, fluid dynamics, optics, thermodynamics, and magnetism. Each scenario is filmed from three camera views, which gives 198 samples, one per scenario and view, and each recording has two real takes. Under the Original protocol, we score 80-frame, five-second continuations at 16 fps. \method receives the event prompt and the switch frame, at which the observed context ends; the recorded continuations are used only for evaluation. The Cosmos3-Nano~\cite{agarwal2026cosmos} and OmniWeaving~\cite{pan2026omniweaving} baselines use their image-to-video (I2V) settings, and Sec.~\ref{sec:supp_omniweaving} describes the OmniWeaving interface used with \method. CoECT~\cite{wang2026chain} is reported both with a generated first frame and with an observed frame from the benchmark. Physics-IQ Verified revises the benchmark data, prompts, and sample-level scoring~\cite{radsch2026verified}.

\paragraph{PhyGenBench.}
\phygen~\cite{meng2024towards} contains 160 prompts covering 27 physical laws, with 40, 50, 30, and 40 prompts in Mechanics, Optics, Thermal, and Material, respectively. The protocol generates one video per prompt in the official order, with a common image source for image-conditioned methods. We report normalized PhyGenEval scores on $[0,1]$, where higher is better.

\paragraph{VBench-2.0.}
\vbench~\cite{zheng2025vbench} evaluates 18 dimensions across human fidelity, controllability, creativity, physics, and commonsense. We report only its physics categories and adopt VideoCoCo's three-category protocol~\cite{li2026videococo}: Mechanics covers mechanical behavior, Thermotics covers thermal state changes, and Material covers properties such as mixing and solubility.

\subsection{Physics-IQ: Masks and Component Metrics}
\label{sec:supp_physics_metrics}

We follow the Original Physics-IQ evaluator~\cite{motamed2026physics}. For scenario $i$, view $v$, and frame $t$, let $\widehat V_{ivt}$ be the generated RGB frame and $V^{(1)}_{ivt},V^{(2)}_{ivt}$ the two real takes. RGB values lie in $[0,1]$. Let $\widehat M_{ivt},M^{(1)}_{ivt},M^{(2)}_{ivt}$ be the binary dynamic-region masks produced by the benchmark preprocessing. Each continuation contains $T=80$ scored frames.

\paragraph{Appearance Error.}
The per-frame RGB mean squared error against the first real take is
\begin{equation}
 E_{ivt}=\frac{1}{3HW}\sum_{x,y,c}
 \left(\widehat V_{ivt}(x,y,c)-V^{(1)}_{ivt}(x,y,c)\right)^2.
 \label{eq:iq_mse_raw}
\end{equation}
Here $H,W$ are the frame dimensions and $c$ indexes the three color channels. This raw error is distinct from the scaled MSE penalty used in the aggregate score below.

\paragraph{Spatiotemporal IoU.}
Framewise mask overlap measures whether motion occupies the correct regions at the correct times:
\begin{equation}
 J^{\mathrm{st}}_{ivt}
 =\frac{|\widehat M_{ivt}\cap M^{(1)}_{ivt}|}
 {|\widehat M_{ivt}\cup M^{(1)}_{ivt}|}.
 \label{eq:iq_st}
\end{equation}
The evaluator assigns one when both masks are empty.

\paragraph{Spatial IoU.}
For the generated video, the union of active regions over the clip is
\begin{equation}
 \widehat U_{iv}(x,y)=\mathbb{I}\!\left[\sum_{t=1}^{T}\widehat M_{ivt}(x,y)>0\right],
\end{equation}
and $U^{(1)}_{iv}$ and $U^{(2)}_{iv}$ are defined in the same way for the two real takes. Spatial overlap then ignores when a region becomes active:
\begin{equation}
 J^{\mathrm{sp}}_{iv}
 =\frac{|\widehat U_{iv}\cap U^{(1)}_{iv}|}
 {|\widehat U_{iv}\cup U^{(1)}_{iv}|}.
 \label{eq:iq_sp}
\end{equation}

\paragraph{Weighted Spatial IoU.}
The temporal occupancy of the generated video at each pixel is
\begin{equation}
 \widehat\omega_{iv}(x,y)=\frac{1}{T}\sum_{t=1}^{T}\widehat M_{ivt}(x,y),
\end{equation}
and $\omega^{(1)}_{iv}$ and $\omega^{(2)}_{iv}$ are defined in the same way for the real takes. Weighted overlap preserves this duration information:
\begin{equation}
 J^{\mathrm{ws}}_{iv}
 =\frac{\sum_{x,y}\min(\widehat\omega_{iv}(x,y),\omega^{(1)}_{iv}(x,y))}
 {\sum_{x,y}\max(\widehat\omega_{iv}(x,y),\omega^{(1)}_{iv}(x,y))}.
 \label{eq:iq_ws}
\end{equation}
Spatial and Weighted Spatial IoU also equal one when the union is empty.

\subsection{Physics-IQ: Original Aggregation}
\label{sec:supp_iq_aggregation}

An overbar denotes averaging over evaluated scenarios and their three views, and over frames for $E$ and $J^{\mathrm{st}}$. Applying the same measurements between the two real takes gives
\begin{equation}
 \nu_E=\overline{E(V^{(1)},V^{(2)})},\qquad
 \nu_\star=\overline{J^\star(M^{(1)},M^{(2)})},
\end{equation}
where $\star\in\{\mathrm{sp},\mathrm{st},\mathrm{ws}\}$. These real-to-real measurements capture natural variation between repeated experiments.

The normalized overlap components and appearance penalty, on the reporting scale used for the Original score, are
\begin{equation}
 \begin{aligned}
 \mathrm{SP}&=100\,\bar J^{\mathrm{sp}}/\nu_{\mathrm{sp}},\\
 \mathrm{ST}&=100\,\bar J^{\mathrm{st}}/\nu_{\mathrm{st}},\\
 \mathrm{WS}&=100\,\bar J^{\mathrm{ws}}/\nu_{\mathrm{ws}},\\
 \mathrm{MSE}&=100\,(\bar E-\nu_E).
 \end{aligned}
 \label{eq:iq_components}
\end{equation}
The aggregate is
\begin{equation}
 S_{\mathrm{IQ}}=
 \operatorname{clip}_{[0,100]}
 \left(\frac{\mathrm{SP}+\mathrm{ST}+\mathrm{WS}}{3}
 -\mathrm{MSE}\right).
 \label{eq:iq_score}
\end{equation}
The released implementation rounds the final result to two decimal places.

The overall calculation uses all 198 scenario-view samples. A domain result recomputes the averages and real-to-real normalizers within that domain, so the overall score is not a scenario-weighted mean of the domain scores.

\subsection{PhyGenEval: Three Evaluation Stages}
\label{sec:supp_phygen_metrics}

PhyGenEval~\cite{meng2024towards} evaluates each prompt $i$ through phenomenon detection, event-order verification, and whole-video naturalness. Each stage produces a score in $\{0,1,2,3\}$.

\paragraph{Stage 1: Phenomenon Detection.}
Let $\operatorname{VQA}(F,q,y)$ denote VQAScore~\cite{lin2024evaluating} for answer $y$ given frame $F$ and question $q$. For a retrieved frame $F_t$ and question $q$, the scores of the positive and negative answers, $\lambda^\pm_{iqt}=\operatorname{VQA}(F_t,q,y^\pm)$, give
\begin{equation}
 \gamma_{iqt}=\frac{\lambda^+_{iqt}}{\lambda^+_{iqt}+\lambda^-_{iqt}},\qquad
 \mu_{iq}=\max_{t\in\mathcal N(t^\star)}(w_{iqt}+\gamma_{iqt}),
\end{equation}
where $w_{iqt}$ is retrieval relevance and $\mathcal N(t^\star)$ is the neighborhood of the selected key frame $t^\star$. Last-frame questions use the final-frame answer score directly.

For a prompt with one question, the stored retrieval contribution is subtracted from $\mu_{i1}$ when retrieval is used. The resulting evidence $\xi_i$ is quantized as
\begin{equation}
 Q_1(\xi)=
 \begin{cases}
 0,&0\le \xi<0.25,\\
 1,&0.25\le \xi<0.50,\\
 2,&0.50\le \xi<0.75,\\
 3,&\xi\ge0.75.
 \end{cases}
\end{equation}
For a prompt with two questions, $\xi_i=\mu_{i1}+\mu_{i2}$ and
\begin{equation}
 Q_2(\xi)=
 \begin{cases}
 0,&\xi\le1,\\
 1,&1<\xi\le1.5,\\
 2,&1.5<\xi\le2.25,\\
 3,&\xi>2.25.
 \end{cases}
\end{equation}
The phenomenon score is $f_i=Q_1(\xi_i)$ or $Q_2(\xi_i)$ according to the question count.

\paragraph{Stage 2: Event Order.}
Let $h=c$ denote the closed track using \gptfouro~\cite{hurst2024gpt} and $h=o$ the open track using \llavainterleave~\cite{li2025llava}. Without retrieval, the score counts affirmative answers to three temporal questions:
\begin{equation}
 m_i^{(h)}=\sum_{q=1}^{3}\mathbb{I}[y_{iq}^{(h)}=\mathrm{Yes}].
\end{equation}
For questions requiring retrieval, define
\begin{equation}
 \beta_{iq}^{(h)}=
 \mathbb{I}\!\left[
 \max_t\{w_{iqt}+\mathbb{I}[y_{iqt}^{(h)}=\mathrm{Yes}]\}
 \ge1.5\right].
\end{equation}
The maximum is over the question's retrieved temporal interval. The closed track uses retrieval for $\mathcal K_c=\{1,2,3\}$ and the open track for $\mathcal K_o=\{1,2\}$:
\begin{equation}
 m_i^{(h)}=\sum_{q\in\mathcal K_h}\beta_{iq}^{(h)}
 +\sum_{q\notin\mathcal K_h}\mathbb{I}[y_{iq}^{(h)}=\mathrm{Yes}].
\end{equation}

\paragraph{Stage 3: Naturalness.}
The whole-video score $n_i^{(h)}$ maps Completely Fantastical, Highly Unrealistic, Slightly Unrealistic, and Almost Realistic to $0,1,2,3$, respectively. \gptfouro supplies the closed-track judgment and InternVideo2~\cite{wang2024internvideo2} the open-track judgment.

\subsection{PhyGenEval: Aggregation}

The official implementation combines the three stage scores separately for the two tracks:
\begin{equation}
 \begin{aligned}
 g_i^{(c)}&=\operatorname{round}\!\left(
 \frac{f_i+m_i^{(c)}+n_i^{(c)}}{3}\right),\\
 g_i^{(o)}&=\operatorname{round}\!\left(
 \frac{f_i+m_i^{(o)}+n_i^{(o)}}{3}\right).
 \end{aligned}
\end{equation}
Here $\operatorname{round}$ follows Python's rounding convention, including ties to the nearest even integer. The combined per-prompt score uses integer floor division:
\begin{equation}
 g_i=\left\lfloor\frac{g_i^{(c)}+g_i^{(o)}}{2}\right\rfloor.
\end{equation}
For a prompt set $\mathcal D$, the normalized automatic score is
\begin{equation}
 S_{\mathrm{PhyGen}}(\mathcal D)
 =\frac{1}{3|\mathcal D|}\sum_{i\in\mathcal D}g_i.
 \label{eq:phygen_score}
\end{equation}
The overall set contains all 160 prompts; domain scores use the corresponding subsets.

\subsection{VBench-2.0 Physics Protocol}
\label{sec:supp_vbench}

\vbench tests physical state changes through visual question answering~\cite{zheng2025vbench}. In the official implementation, a video that fails an initial-state question is excluded from that category's denominator. The remaining videos receive binary judgments on the expected visible phenomenon. Questions without an initial-state filter are scored directly. This evaluator filtering is separate from \method's routing and fallback.

For category $c\in\{\mathrm{Mechanics},\mathrm{Thermotics},\mathrm{Material}\}$, let $\mathcal V_c$ be the evaluator-valid videos and $y_{c,i}\in\{0,1\}$ the phenomenon judgment for video $i$. The percentage score and the three-category mean are
\begin{equation}
 S_c=\frac{100}{|\mathcal V_c|}\sum_{i\in\mathcal V_c}y_{c,i},
 \qquad S_{\mathrm{AVG}}=\frac{1}{3}\sum_c S_c.
 \label{eq:vbench_score}
\end{equation}
For nonempty $\mathcal V_c$, this follows the official category aggregation and VideoCoCo's three-category reporting convention~\cite{li2026videococo}. The mean weights the three categories equally regardless of their numbers of valid videos and leaves out Multi-View Consistency and the other benchmark dimensions.

With \method, OmniWeaving scores 81.41, 65.54, and 81.54 on Mechanics, Thermotics, and Material, which average to 76.16. The six comparison rows of Tab.~\mainfloatref{tab:vbench}\inmainpaper{} come from Tab.~2 of VideoCoCo~\cite{li2026videococo}.

\section{OmniWeaving Configuration}
\label{sec:supp_omniweaving}

OmniWeaving supports joint text, image, and video conditioning~\cite{pan2026omniweaving}. With \method, we use its \texttt{tiv2v} mode, which receives the expanded event prompt, the Physics-IQ switch frame, and the depth control of \method as a video input. The prompt instructs the generator to take motion geometry and event timing from the depth video while producing a realistic RGB video.

With \method, the Physics-IQ score of OmniWeaving rises from 29.69 to 33.71 under the Original protocol and from 28.64 to 33.21 on Physics-IQ Verified~\cite{radsch2026verified} (Tab.~\mainfloatref{tab:physicsiq}\inmainpaper).

\section{Additional Ablation Studies}
\label{sec:supp_ablations}

These studies use frozen \wanvace and the Original Physics-IQ score, and each row is a complete configuration.

\subsection{System Components}
\label{sec:ablation_components}

This study asks whether simulated depth alone accounts for the improvement. Starting from \wanvace with the expanded prompt $\widetilde p$, we add simulated depth without checks, program validation and capability checks, rollout verification, and finally repair with capability routing. Untyped simulation builds an executable simulation from an unconstrained scene description and renders its first executable rollout without checks. Typed simulation adds the typed program and capability checks but does not verify the rollout. Verified simulation verifies the rollout but does not modify a failed program. Full \method adds repair and sends inputs that fail capability checks or never pass to the fallback. All variants use the same generation settings and return one output per sample.

\begin{table*}[!t]
  \caption{\textbf{System components on Physics-IQ.} All variants use frozen \wanvace~\cite{wan2025wan,jiang2025vace} and the Original score. Red and blue mark the best and second-best values.}
  \label{tab:supp_components}
  \label{tab:ablation_components}
  \centering
  \small
  \begin{tabular}{@{}lcccccc@{}}
    \toprule
    Variant & Solid $\uparrow$ & Fluid $\uparrow$ & Optics $\uparrow$ & Thermodynamics $\uparrow$ & Magnetism $\uparrow$ & Overall $\uparrow$ \\
    \midrule
    \wanvace~\cite{wan2025wan,jiang2025vace} & 30.42 & 24.23 & 28.07 & 14.34 & 12.56 & 27.33 \\
    Untyped simulation & 34.86 & 29.74 & 28.42 & 14.72 & 13.48 & 31.20 \\
    Typed simulation & 38.12 & 33.67 & 29.18 & 15.56 & 16.72 & 34.26 \\
    Verified simulation & \second{41.22} & \second{37.42} & \second{30.36} & \second{16.83} & \second{19.68} & \second{37.21} \\
    Full \method & \best{43.46} & \best{39.94} & \best{31.27} & \best{18.22} & \best{22.65} & \best{39.40} \\
    \bottomrule
  \end{tabular}
\end{table*}

Each added stage raises the score in every domain (Tab.~\ref{tab:ablation_components}).

\subsection{Verification Evidence}
\label{sec:supp_ablation_evidence}

This study compares VLM judgments, physical measurements, and their combination. The VLM checks only identity, expected effect, and event order. Physical measurements assess the process and the final state, and a separate control check measures visibility in depth control. All variants share the generator, routing, and rollout budget.

\begin{table*}[!t]
  \caption{\textbf{Verification evidence by domain on Physics-IQ.} Original scores with frozen \wanvace~\cite{wan2025wan,jiang2025vace} and the same rollout budget. The hybrid combines VLM judgments with physical measurements. Red and blue mark the best and second-best values.}
  \label{tab:supp_evidence}
  \centering
  \small
  \begin{tabular}{@{}lcccccc@{}}
    \toprule
    Variant & Solid $\uparrow$ & Fluid $\uparrow$ & Optics $\uparrow$ & Thermodynamics $\uparrow$ & Magnetism $\uparrow$ & Overall $\uparrow$ \\
    \midrule
    VLM only & 39.72 & 35.66 & \second{29.84} & \second{17.61} & \second{20.44} & 36.00 \\
    Measurements only & \second{41.38} & \second{38.21} & 28.72 & 16.48 & 19.36 & \second{37.28} \\
    Hybrid verifier & \best{43.46} & \best{39.94} & \best{31.27} & \best{18.22} & \best{22.65} & \best{39.40} \\
    \bottomrule
  \end{tabular}
\end{table*}

The hybrid verifier scores highest in every domain and overall. Physical measurements score higher than VLM judgments in solids and fluids, and VLM judgments score higher in optics, thermodynamics, and magnetism, so the two sources complement each other.

\subsection{Repair and Capability Routing}
\label{sec:supp_ablation_repair}

This study asks how the response to a failed check affects the final output. All variants use the same Physics-IQ samples, rollout budget, and maximum number of repair attempts. No repair rejects a failed rollout without editing its program, global rebuild recreates the whole program, and localized repair edits only the fields $\Omega_j$ of the failed check. These three variants always use simulation and return simulated depth even if no rollout passes. Full routing uses localized repair and, as in Alg.~\ref{alg:physmend}, sends inputs that fail capability checks or never pass to the fallback.

\begin{table*}[!t]
  \caption{\textbf{Repair and routing by domain on Physics-IQ.} The first three variants always use simulation; full routing permits the fallback. Red and blue mark the best and second-best values.}
  \label{tab:supp_repair}
  \centering
  \small
  \begin{tabular}{@{}lcccccc@{}}
    \toprule
    Variant & Solid $\uparrow$ & Fluid $\uparrow$ & Optics $\uparrow$ & Thermodynamics $\uparrow$ & Magnetism $\uparrow$ & Overall $\uparrow$ \\
    \midrule
    No repair & 41.22 & 37.42 & 30.36 & 16.83 & 19.68 & 37.21 \\
    Global rebuild & 41.86 & 38.08 & 30.48 & 16.97 & 20.12 & 37.78 \\
    Localized repair & \second{42.82} & \second{39.18} & \second{30.76} & \second{17.24} & \second{20.84} & \second{38.66} \\
    Full routing & \best{43.46} & \best{39.94} & \best{31.27} & \best{18.22} & \best{22.65} & \best{39.40} \\
    \bottomrule
  \end{tabular}
\end{table*}

Localized repair scores above global rebuild in every domain with the same budgets, and full routing adds further gains, most in magnetism and thermodynamics.

\subsection{Control Representation}
\label{sec:ablation_control}

To examine how the motion representation affects generation, we compare image and text, coarse guidance, raw simulation depth, and verified dense depth with a shared \wanvace generator. Coarse guidance uses sparse keyframes or projected object trajectories without dense spatial control. Raw simulation depth renders the simulated scene to depth without visibility correction, background reconstruction, normalization, smoothing, or control checks. Verified dense depth is full \method: it uses the full control construction, and only rollouts that pass all checks send their depth control to the generator.

\begin{table*}[!t]
  \caption{\textbf{Control representation on Physics-IQ.} Original scores with frozen \wanvace~\cite{wan2025wan,jiang2025vace}. Verified dense depth is full \method. Red and blue mark the best and second-best values.}
  \label{tab:supp_control}
  \label{tab:ablation_control}
  \centering
  \small
  \begin{tabular}{@{}lcccccc@{}}
    \toprule
    Variant & Solid $\uparrow$ & Fluid $\uparrow$ & Optics $\uparrow$ & Thermodynamics $\uparrow$ & Magnetism $\uparrow$ & Overall $\uparrow$ \\
    \midrule
    Image and text & 30.42 & 24.23 & 28.07 & 14.34 & 12.56 & 27.33 \\
    Coarse guidance & 34.95 & 29.08 & 28.44 & 14.62 & 13.72 & 31.17 \\
    Raw simulation depth & \second{39.62} & \second{35.89} & \second{29.76} & \second{16.11} & \second{17.64} & \second{35.72} \\
    Verified dense depth & \best{43.46} & \best{39.94} & \best{31.27} & \best{18.22} & \best{22.65} & \best{39.40} \\
    \bottomrule
  \end{tabular}
\end{table*}

Raw simulation depth scores above coarse guidance, and verified dense depth above raw depth, in every domain (Tab.~\ref{tab:ablation_control}).

\section{Generation Calls and Computational Cost}
\label{sec:supp_efficiency}

\paragraph{Computation Before Synthesis.}
A generation call invokes the video generator to synthesize one RGB clip. Simulation and verification are separate operations: the default algorithm allows one initial rollout and up to three correction rollouts, but invokes the generator only after accepting control or choosing image-and-prompt fallback. A rejected program revision therefore costs another simulation and verification pass without another RGB synthesis.

\begin{figure*}[!t]
  \centering
  \includegraphics[width=\textwidth]{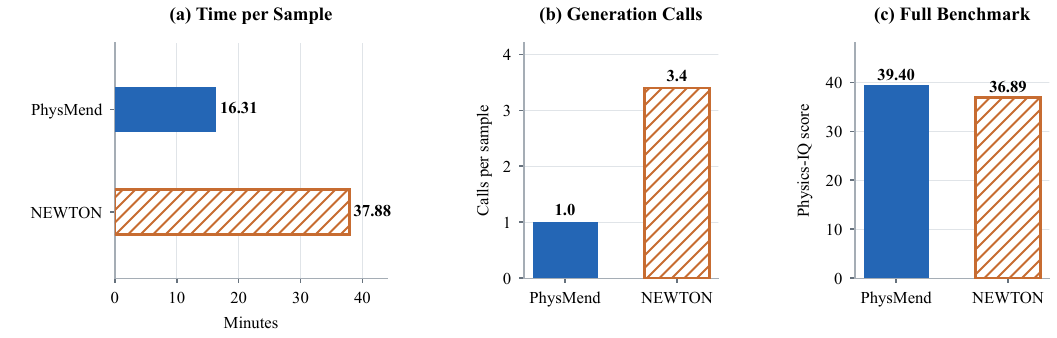}
  \caption{\textbf{Generation calls, time, and Physics-IQ score of \method (blue) and NEWTON~\cite{feng2026newton} (hatched).} (a) Mean time per sample on the full Physics-IQ benchmark. For \method, it is the recorded pipeline time with inference on a single 80 GB H100 GPU; for NEWTON, it is a generation-only estimate, 3.4 calls times the measured time of one generation call. (b) Mean RGB generation calls per sample over the full Physics-IQ benchmark, including NEWTON's regenerations. (c) Overall Physics-IQ score on the full benchmark.}
  \label{fig:generation_calls}
\end{figure*}

\begin{figure*}[!t]
  \centering
  \includegraphics[width=\textwidth]{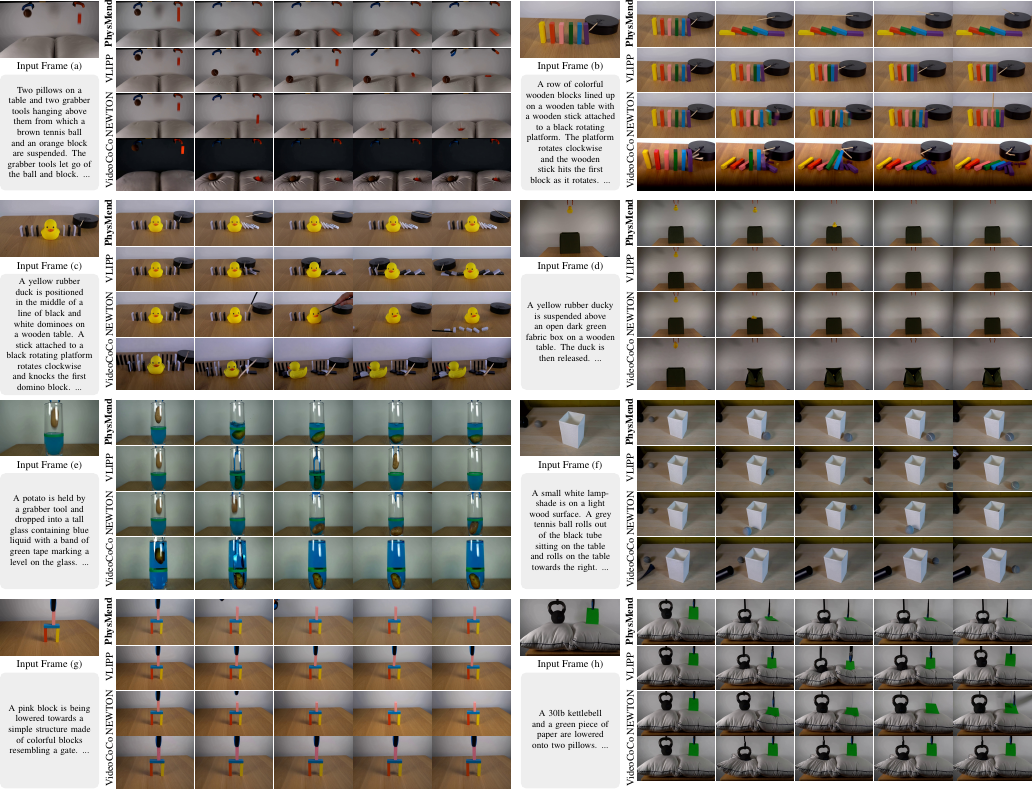}
  \caption{\textbf{Additional qualitative comparisons on Physics-IQ.} We compare \method with VLIPP~\cite{yang2025vlipp}, NEWTON~\cite{feng2026newton}, and VideoCoCo~\cite{li2026videococo} on eight center-view scenes from \physicsiq~\cite{motamed2026physics}: (a) a ball and a block released above pillows, (b) a rotating stick hitting a row of blocks, (c) a line of dominoes with a duck in the middle, (d) a duck released above a box, (e) a potato dropped into liquid, (f) a ball rolling behind a lampshade, (g) a block lowered onto a gate, and (h) a kettlebell and a sheet of paper lowered onto pillows. Each panel shows the benchmark's switch frame, an excerpt of its original prompt, and five chronological frames per method. Ellipses indicate omitted prompt text. \method occupies the first row. Frames keep their original aspect ratios.}
  \label{fig:qualitative_additional}
\end{figure*}

\paragraph{Comparison with NEWTON.}
NEWTON~\cite{feng2026newton} can generate another video after planning or verification, whereas \method revises the program and its depth control before one final generation call. \method therefore makes one RGB generation call per sample. Averaged over the full Physics-IQ benchmark, NEWTON makes 3.4 generation calls per sample, including its regenerations, so \method uses 70.6\% fewer generation calls (Fig.~\ref{fig:generation_calls}(b)).

\paragraph{Recorded Time.}
The recorded pipeline of \method covers simulation, preparation, and generation. On the full Physics-IQ benchmark, it takes 16.31 minutes per sample on average, of which simulation takes 5.11 minutes. The single generation call takes 9.58 to 11.60 minutes per sample, so differences in pipeline time come mainly from simulation. For NEWTON, 3.4 calls times the mean measured time of one generation call give 37.88 minutes (Fig.~\ref{fig:generation_calls}(a)); this estimate counts generation only, so NEWTON's full runtime is longer. On the full benchmark, \method scores 39.40 on Physics-IQ and NEWTON 36.89 (Fig.~\ref{fig:generation_calls}(c)).

\section{Motion-Conditioned Generation with Seedance 2.5}
\label{sec:supp_seedance}

\paragraph{Purpose and Setup.}
\method passes simulated motion to the generator as a depth video. We test whether such motion guidance also helps Seedance 2.5~\cite{seed2026one}, which receives it through a reference-video interface instead of the dense-control interface of \wanvace. This study uses a ten-scene subset of Physics-IQ~\cite{motamed2026physics}. Its I2V baseline receives the switch frame and the event prompt through the \texttt{image-to-video} endpoint. Three reference-to-video (R2V) conditions keep the same switch frame and add a particle-motion, coarse-RGB, or depth video through the \texttt{reference-to-video} endpoint, so they show the same event motion in different visual forms.

\begin{table}[!htbp]
  \caption{\textbf{Motion guidance for Seedance 2.5.} All rows use the same ten-scene Physics-IQ subset. I2V uses the switch frame and prompt; R2V adds a motion video. Scores are on the percentage scale, and gains are percentage points relative to I2V. Red and blue mark the best and second-best values.}
  \label{tab:seedance}
  \centering
  \small
  \begin{tabular}{@{}llcc@{}}
    \toprule
    Mode & Motion Video & Score $\uparrow$ & Gain (pp) \\
    \midrule
    I2V & None & 41.29 & 0.00 \\
    R2V & Particle motion & 45.69 & +4.40 \\
    R2V & Coarse RGB & \second{46.08} & \second{+4.79} \\
    R2V & Depth & \best{49.66} & \best{+8.37} \\
    \bottomrule
  \end{tabular}
\end{table}

\paragraph{Results.}
All three motion videos improve on I2V, showing that explicit motion guidance remains useful for Seedance 2.5 (Tab.~\ref{tab:seedance}). Depth provides the largest gain, raising the Physics-IQ score from 41.29 to 49.66, an improvement of 8.37 percentage points. Its Spatiotemporal IoU also increases from 0.1688 to 0.3106 on the raw overlap scale, so the generated motion matches the recording better in both location and timing. Particle and coarse-RGB videos also help, but neither matches the depth condition, so the gain depends on how the motion is represented, not only on supplying another video.
\section{Additional Evaluation Settings}
\label{sec:supp_reporting}

\paragraph{Implementation and Baselines.}
\label{sec:supp_implementation}
\label{sec:settings}
The default generator is frozen \wanvace~\cite{wan2025wan,jiang2025vace}. As a baseline, \wanvace alone receives $I_0$ and the expanded prompt $\widetilde p$ without depth control. To test other control interfaces, we also pair \method with Cosmos3-Nano~\cite{agarwal2026cosmos}, which takes the depth control through its built-in depth conditioning without additional LoRA weights~\cite{hu2021lora}, and with OmniWeaving~\cite{pan2026omniweaving}, which takes it as a video input in its unified \texttt{tiv2v} mode. We compare with direct generators~\cite{yang2025cogvideox,hacohen2024ltx,blattmann2023stable}, prompt and event planners~\cite{xue2025phyt2v,wang2026chain,yang2025vlipp,huang2025planning}, motion-control methods~\cite{namekata2025sg,burgert2025go}, and agentic workflows~\cite{feng2026newton,li2026videococo}, and run NEWTON and SketchVerify on \physicsiq with their public code.

\paragraph{Generation Settings and Reproducibility.}
The \wanvace configuration uses $832\times480$, 81 frames, 16 fps, 50 inference steps, guidance 5.0, context scale 1.0, and deterministic scenario seeds. Its checkpoint is named both Wan2.2-VACE and Wan2.2-VACE-Fun-A14B~\cite{wan2025wan}, and \wanvace denotes this configuration. All verification and repair hyperparameters are shared across benchmarks, and full-benchmark scores include inputs that used the fallback.

\section{Workflow Example}
\label{sec:supp_figures}

The overview figure (Fig.~\mainfloatref{fig:overview}\inmainpaper) uses the ball-in-sand example to show the observed image, object overlay, particle states, depth control, and generated RGB frames. It shows horizontal center crops of these frames, and its program follows the recorded program in Tab.~\ref{tab:supp_program_example}. Blue, green, and gray denote the ball, sand, and fixed scene, respectively. This rollout passed every check on the first attempt, so the figure follows a granular interaction through the pipeline without repair.

The verification figure (Fig.~\mainfloatref{fig:verification}\inmainpaper) follows one representative \physicsiq{} sample in which a ball and a block fall onto two pillows. Initially, the program labels the hanging block a static fixture, and the solver uses the compiler's default substeps instead of its setting for stiff rigid--deformable impacts. Each rollout is a full simulation that the VLM and the physical measurements check, and the edits follow the rules of Sec.~\ref{sec:supp_repair}: the failed numerical-health check multiplies the substeps by 1.5, and the failed release of an object labeled fixed triggers replanning of its role. The simulation column shows each rollout rendered as particles in the input camera.

\section{Additional Qualitative Results}
\label{sec:supp_qualitative}

Fig.~\ref{fig:qualitative_additional} compares \method with VLIPP~\cite{yang2025vlipp}, NEWTON~\cite{feng2026newton}, and VideoCoCo~\cite{li2026videococo} on eight more center-view scenes from \physicsiq~\cite{motamed2026physics}, using the methods, frame sampling, and layout of Fig.~\mainfloatref{fig:qualitative}\inmainpaper. Each panel shows the benchmark's switch frame and an excerpt of its original prompt.

With \method, the stick topples the whole row of blocks (b), only the dominoes between the platform and the duck fall (c), the potato sinks into the liquid (e), the grabber releases the pink block, which stays upright on the gate (g), and the paper settles flat on one pillow while the kettlebell rests on the other (h). In contrast, the blocks in (b) stay upright with VLIPP and NEWTON, a hand holding a stick appears with NEWTON in (c), the potato rises back out of the liquid with VLIPP in (e), and the grabber keeps holding the pink block in (g) with VLIPP and VideoCoCo.

\section{Limitations}
\label{sec:limitations}

\method builds its simulation from estimated masks and relative depth, so occlusion, transparency, and hidden surfaces can affect the simulated contacts. Its coverage is bounded by the materials, interactions, and checks that the simulator and verifier support, and depth control carries geometry and motion but not temperature, phase, or optical state. Extending the simulator, the check set, and the control signals is a main direction for future work.

\ifdefined\standalonesupplement
  \setbox0=\vbox{%
  \InputIfFileExists{\fullbblfile}{}{%
    \InputIfFileExists{main.bbl}{}{%
      \PackageError{PhysMend}{Missing \fullbblfile}%
        {Build main.tex into output/cvpr first, e.g. with scripts/build_cvpr_split.sh.}}}}
\fi
\fi 
\end{document}